\documentclass[11pt,a4paper]{article}
\usepackage[hyperref]{eacl2021}
\usepackage{times}
\usepackage{latexsym}

\usepackage{microtype}

\aclfinalcopy 
\def\aclpaperid{478} 

\usepackage{booktabs}
\usepackage{dsfont}
\usepackage{amssymb}
\usepackage{amsmath}
\usepackage[nameinlink,capitalise]{cleveref}
\usepackage[inline]{enumitem}
\usepackage{interval}
\usepackage{graphicx}
\usepackage{subcaption}
\usepackage{tabularx}
\usepackage{multirow}

\usepackage[whole]{bxcjkjatype}

\crefformat{section}{\S #2#1#3}
\Crefformat{section}{\S #2#1#3}

\newcommand{\para}[1]{\noindent\textbf{#1.}}
\newcommand{\entails}[2]{\textit{#1} $\Rightarrow$ \textit{#2}}
\newcommand{\type}[1]{\textsc{#1}}
\newcommand{\dev}[1]{dev\textsubscript{#1}}
\newcommand{\R}{\mathds{R}}  
\newcommand{\pat}{\Phi}
\newcommand{\apat}{\Psi} 
\newcommand{\p}{\varphi}
\newcommand{\anti}{\psi}
\newcommand{\thr}{\vartheta}
\newcommand{\abs}[1]{\left|#1\right|}
\newcommand{\eset}[1]{\left\{\, #1 \,\right\}}
\newcommand{\spectok}[1]{\mathtt{\langle #1 \rangle}}
\newcommand{\correct}[1]{{\textcolor{green}{#1}}}
\newcommand{\wrong}[1]{\underline{\textcolor{red}{#1}}}
\newcommand{\lm}{\Lambda}
\newcommand{\predictions}[6]{\multicolumn{1}{r}{truth: #1 \textsc{nli}: #2 \textsc{manpat}: #3 / #4 \textsc{autpat}: #5 / #6}}

\title{Language Models for Lexical Inference in Context}

\author{Martin Schmitt\textsuperscript{1} \and Hinrich Schütze\textsuperscript{2} \\
	Center for Information and Language Processing (CIS)\\
	LMU Munich, Germany\\
	\textsuperscript{1}\texttt{martin@cis.lmu.de} \quad \textsuperscript{2}\texttt{inquiries@cislmu.org}\\}

\date{}

\begin{document}
\maketitle
\begin{abstract}
Lexical inference in context (LIiC) is the task of recognizing textual entailment
between two very similar sentences, i.e., sentences that only differ in one expression.
It can therefore be seen as a variant of the natural language inference task
that is focused on lexical semantics.
We formulate and evaluate the first approaches based on pretrained language models (LMs) for this task:
\begin{enumerate*}[label=(\roman*)]
	\item a few-shot NLI classifier,
	\item a relation induction approach based on handcrafted patterns expressing the semantics of lexical inference, and
	\item a variant of (ii) with patterns that were automatically extracted from a corpus.
\end{enumerate*}
All our approaches outperform the previous state of the art,
showing the potential of pretrained LMs for LIiC.
In an extensive analysis, we investigate factors of success and failure of our three approaches.\footnote{Our code is publicly available: \url{https://github.com/mnschmit/lm-lexical-inference}}
\end{abstract}

\section{Introduction}
Lexical inference (LI) denotes the task of deciding
whether or not an entailment relation holds between two lexical items.
It is therefore related to the detection of other lexical relations like hyponymy between nouns \citep{hearst-1992-automatic}, e.g., \entails{dog}{animal}, or troponymy between verbs \citep{fellbaum90}, e.g., \entails{to traipse}{to walk}.
Lexical inference in context (LIiC) adds the problem of disambiguating the pair of lexical items in a given context before reasoning about the inference question.
This type of LI is particularly interesting for entailments between verbs and verbal expressions
because their meaning
-- and therefore their implications --
can drastically change with different arguments.
Consider, e.g., \entails{run}{lead} in a \type{person} / \type{company} context (``Bezos runs Amazon'') vs.\ \entails{run}{execute} in a \type{computer} / \type{software} context (``My mac runs macOS'').
LIiC is thus also closely related to the task of natural language inference (NLI)
-- also called recognizing textual entailment \citep{dagan13} --
and can be seen as a focused variant of it.
Besides the important use case of evaluating NLI systems,
this kind of predicate entailment has also been shown useful for question answering \citep{schoenmackers-etal-2010-learning},
event coreference \citep{shwartz-etal-2017-acquiring,meged20},
and link prediction in knowledge graphs \citep{hosseini-etal-2019-duality}.

Despite its NLI nature,
previous systems for LIiC have primarily been models of lexical similarity \citep{levy-dagan-2016-annotating} or models based on verb argument inclusion \citep{hosseini-etal-2019-duality}.
The reason is probably that supervised NLI models need large amounts of training data, which is unavailable for LIiC, and that systems trained on available large-scale NLI benchmarks \citep[e.g.,][]{williams-etal-2018-broad} have been reported to insufficiently cover lexical phenomena \citep{glockner-etal-2018-breaking,schmitt-schutze-2019-sherliic}.

Recently, transfer learning has become ubiquitous in NLP;
Transformer \citep{vaswani17} language models (LMs) pretrained on large amounts of textual data \citep{devlin19,liu19}
form the basis of a lot of current state-of-the-art models.
Besides zero- and few-shot capabilities \citep{radford19,brown20},
pretrained LMs have also been found to acquire factual and relational knowledge during pretraining \citep{petroni-etal-2019-language,bouraoui20}.
The entailment relation certainly stands out among previously explored semantic relations
-- such as the relation between a country and its capital --
because it is very rarely stated explicitly and often involves reasoning about both the meaning of verbs and additional knowledge \citep{schmitt-schutze-2019-sherliic}.
It is unclear
whether implicit clues during pretraining are enough to learn about LIiC
and what the best way is to harness any such implicit knowledge.

Regarding these questions,
we make the following contributions:
\begin{enumerate*}[label=(\arabic*)]
	\item This work is the first to explore the use of pretrained LMs for the LIiC task.
	\item We formulate three approaches and evaluate them using the publicly available pretrained RoBERTa LM \citep{liu19,huggingface}:
		\begin{enumerate*}[label=(\roman*)]
			\item a few-shot NLI classifier,
			\item a relation induction approach based on handcrafted patterns expressing the semantics of lexical inference, and
			\item a variant of (ii) with patterns that were automatically extracted from a corpus.
		\end{enumerate*}
	\item We introduce the concept of antipatterns, patterns that express non-entailment, and evaluate their usefulness for LIiC.
	\item In our experiments on two established LIiC benchmarks,
	Levy/Holt's dataset \citep{levy-dagan-2016-annotating,holt18} and SherLIiC \citep{schmitt-schutze-2019-sherliic},
	all our approaches consistently outperform previous work,
	thus setting a new state of the art for LIiC.
	\item In contrast to previous work on relation induction \citep{bouraoui20},
	automatically retrieved patterns do not outperform handcrafted ones for LIiC.
	A qualitative analysis of  patterns and errors 
	identifies possible reasons for this finding.
\end{enumerate*}

\section{Related Work}

\para{Lexical inference}
There has been a lot of work on lexical inference for nouns,
notably hypernymy detection,
resulting in a variety of benchmarks \citep{kotlerman10,kiela-etal-2015-exploiting}
and methods \citep{shwartz-etal-2015-learning,vulic-mrksic-2018-specialising}.
Although there has been work on predicate entailment before \citep{lin01,lewis-steedman-2013-combined},
\citet{levy-dagan-2016-annotating} were the first to create a general benchmark for evaluating entailment between verbs.
In their evaluation, neither resource-based approaches \citep{pavlick-etal-2015-ppdb,berant-etal-2011-global}
nor vector space models \citep{levy-goldberg-2014-dependency} achieved satisfying results.
\citet{holt18} later published a re-annotated version, which was readily adopted by later work.
\citet{hosseini-etal-2018-learning} put global constraints on top of directed local similarity scores \citep{weeds-weir-2003-general,lin-1998-automatic-retrieval,szpektor-dagan-2008-learning}
based on distributional features of the predicates.
\citet{hosseini-etal-2019-duality} replaced these scores by transition probabilities in a bipartite graph
where edge weights are computed by a link prediction model.

When \citet{schmitt-schutze-2019-sherliic} created the SherLIiC benchmark,
they also mainly focused on resource- and vector-based models for evaluation.
Their best model combines general-purpose word2vec representations \citep{mikolov13}
with a vector representation of the arguments that co-occur with a predicate.

All these works
\begin{enumerate*}[label=(\roman*)]
	\item base the probability of entailment validity on the similarity of the verbs and
	\item compute this similarity via (expected) co-occurrence of verbs and their arguments.
\end{enumerate*}
Our work differs in that our models solely reason about the sentence surface in an end-to-end NLI task
without access to previously observed argument pairs.
This is possible because our models have learned about these surface forms during pretraining.

\para{Patterns and entailment}
Pattern-based approaches have long been known for hypernymy detection \citep{hearst-1992-automatic}.
Recent work combined them with vector space models \citep{mirkin-etal-2006-integrating,roller-erk-2016-relations,roller-etal-2018-hearst}.
While there are effective patterns, such as \textit{$X$ is a $Y$}, that are indicative for entailment between nouns,
there is little work on comparable patterns for verbs.
\citet{schwartz-etal-2015-symmetric} mine symmetric patterns for lexical similarity
and achieve good results for verbs.
Entailment, however, is not symmetric.

\citet{chklovski-pantel-2004-verbocean} handcrafted 35 patterns
to distinguish 6 semantic relations for pairs of distributionally similar verbs.
Some of their classes like strength (\textit{taint :: poison}) or antonymy (\textit{ban :: allow})
can be indicators of entailment and non-entailment
but are, in general, much more narrowly defined than the patterns we use in our approach.
Another difference to our work is that verb pairs are scored based on co-occurrence counts on the web,
while we employ an LM,
which does not depend on a valid entailment pair actually appearing together in a document.

\para{Patterns and language models}
\citet{amrami-goldberg-2018-word} were the first to manipulate LM predictions
with a simple pattern to enhance the quality of substitute words in a given context
for word sense induction.
\citet{petroni-etal-2019-language} found that large pretrained LMs can be queried for factual knowledge,
when presented with appropriate pattern-generated cloze-style sentences.
This zero-shot factual knowledge has later been shown to be quite fragile \citep{kassner-schutze-2020-negated}.
So we rather focus on approaches that fine-tune an LM on at least a few samples.
\citet{forbes19} train a binary classifier on top of a fine-tuned BERT \citep{devlin19} to predict the truth value of
handwritten statements about objects and their
properties.
While their experiments investigate BERT's physical common sense reasoning,
we focus on the different phenomenon of entailment between two actions expressed by verbs in context.

\citet{schick20a} used handcrafted patterns and LMs for few-shot text classification.
Based on manually defined label-token correspondences,
the predicted classification label is determined by the
token an LM estimates as most probable
at a masked position in the cloze-style pattern.
We differentiate entailment and non-entailment
via compatibility scores for patterns and antipatterns
and not via different predicted tokens.

Addressing relation induction, \citet{bouraoui20} propose an
automatic way of finding, given a relation, LM patterns
that are likely to express it.
They train a binary classifier per relation on the sentences
generated by these patterns.
While some of the relations they consider are related to verbal entailment (e.g., \textit{cook activity-goal eat}),
most of them concern common sense (e.g., \textit{library location-activity reading})
or encyclopedic knowledge (e.g., \textit{Paris capital-of France}).
We adapt their method for the automatic retrieval of promising patterns for LIiC,
but find that handcrafted patterns that capture the generality of the entailment relation
still have an advantage over automatic patterns for LIiC.
Another important novelty we introduce is the use of antipatterns.
While \citet{bouraoui20} have to use negative samples for training their classifiers,
they only consider patterns that exemplify the desired relation.
In contrast, we also use antipatterns that exemplify what the entailment relation is \textbf{not}.
We believe that antipatterns are particularly useful for entailment detection
because they can help identify other kinds of semantic relations
that often pose a challenge to vector space models \citep{levy-dagan-2016-annotating,schmitt-schutze-2019-sherliic}.

\section{Proposed Approaches}

\subsection{NLI classifier}

Building an NLI classifier on top of a pretrained LM
usually means
taking an aggregate sequence representation of the concatenated premise and hypothesis
as input features of a neural network classifier \citep{devlin-etal-2019-bert}.
For RoBERTa \citep{liu19}, this representation is the final hidden state of a special $\spectok{s}$ token
that is prepended to the input sentences,
which in turn are separated by a separator token $\spectok{/s}$.
Let $\lm$ be the function that maps such an input $x = x_1 \spectok{/s} x_2$
to the aggregate representation $\lm(x) \in \R^d$.
Following \citep{devlin-etal-2019-bert,liu19},
we then feed these features to a 2-layer feed-forward neural network with tanh activation:
\begin{align}
	\label{eq:nli_prob}
	\begin{split}
		h(x) &= \mathrm{tanh}(\mathrm{drop}(\lm(x)) W_1 + b_1)\\
		P_{\textsc{nli}}(y \mid x) &= \sigma(\mathrm{drop}(h(x)) W_2 + b_2)
	\end{split}
\end{align}
where $\mathrm{drop}$ applies dropout with a probability of $0.1$,
$\sigma$ is the softmax function,
and $W_1 \in \R^{d\times d}, W_2 \in \R^{d\times 2}, b_1\in\R^d, b_2\in\R^2$ are learnable parameters.
Note that $W_1$ and $b_1$ are still part of the LM's pretrained parameters; so we only train $W_2$ and $b_2$ from scratch.\footnote{We follow the official implementation; cf.\ Jacob Devlin's comment on issue 43 in the BERT GitHub repository, \url{https://github.com/google-research/bert/issues/43}, (accessed 19 January 2021).}
The actual classification decision uses a threshold $\thr$:
\begin{equation*}
	D_{\textsc{nli}}^\thr(x_1, x_2) = \begin{cases}
		1, & \text{if $P_{\textsc{nli}}(y=1\,|\,x_1, x_2) > \thr$}\\
		0, & \text{otherwise}
	\end{cases}
\end{equation*}
The traditional choice for the threshold is $\thr = 0.5$
because that means $D_{\textsc{nli}}^\thr(x_1, x_2) = 1$ iff $P_{\textsc{nli}}(y=1 \mid x_1, x_2) > P_{\textsc{nli}}(y=0 \mid x_1, x_2)$.
We nevertheless keep $\thr$ as a hyperparameter to be tuned on held-out development data.

We train the \textsc{nli} approach by minimizing the negative log-likelihood $\mathcal{L}_{\textsc{nli}}$ of the training data $\mathcal{T}$:
\begin{equation*}
	\mathcal{L}_{\textsc{nli}}(\mathcal{T}) = \sum_{(x_1, x_2, y)\in\mathcal{T}} -\log(P_{\textsc{nli}}(y \mid x_1, x_2))
\end{equation*}

\subsection{Pattern-based classifier}

This approach puts the input sentences $x_1, x_2$ together in a pattern-based textual context and
trains a classifier to distinguish between
felicitous
and
infelicitous
utterances.\footnote{\citet{bouraoui20} called this natural vs.\ unusual.}
In contrast to previous approaches \citep{forbes19,bouraoui20},
we also consider antipatterns that exemplify what kind of semantic relatedness we are not interested in,
and combine probabilities for patterns and antipatterns in the final classification.

\para{Finding suitable patterns}
A simple handcrafted pattern to check for the validity of an inference $x_1\Rightarrow x_2$
is ``$x_2$ \textit{because} $x_1$.''.
An analoguos antipattern  is ``\textit{It is not sure that} $x_2$ \textit{just because} $x_1$.''.
Based on similar considerations,
we manually design 5 patterns and 5 antipatterns
(see \cref{tab:pattern_comparison}).
We will refer to the approach using these handcrafted patterns as \textsc{manpat}.

\citet{bouraoui20} argue that text produced by simple, handcrafted patterns is artificial
and therefore suboptimal for LMs  pretrained on
naturally occurring text.
To adapt their setup to verbal expressions used in LIiC,
we identify suitable patterns (antipatterns) by searching a large text corpus\footnote{We use the Wikipedia dump from Jan 15th 2011.}
for sentences that contain both elements of valid (invalid) entailment pairs.
In a second step, we score each of these patterns (antipatterns) according to the number of valid (invalid) entailment pairs $x_1, x_2$
that can be found by querying an LM for the $k$ most probable completions
when $x_1$ or $x_2$ is inserted in the pattern and its counterpart is masked.
For example, consider the entailment pair \entails{rule}{control} and the pattern ``\textit{Catchers \textbf{prem} the field; they \textbf{hypo} the plays and tell everyone where to be.}'' extracted from a description of softball.
Predicting \textit{rule} from ``\textit{Catchers $\spectok{mask}$ the  field; they control the plays and tell everyone where to be.}''
and predicting \textit{control} from ``\textit{Catchers rule the  field; they $\spectok{mask}$ the plays and tell everyone where to be.}''
would  result in one point each.
Approaches called \textsc{autpat}\textsubscript{$n$} use the $n$ patterns with the most points obtained in that manner.
See \cref{sec:experiments} for
more  details on our experimental setup.

\para{Pattern-based predictions}
The probability $P_{\textsc{fel}}(z \mid x)$ of 
sentence $x$ to be
felicitous
($z\!=\!1$) or
infelicitous
($z\!=\!0$)
is estimated like $P_{\textsc{nli}}$
in \cref{eq:nli_prob}, except
that $x$ is not the concatenation of two sentences
but a single pattern-generated utterance.

Given a set of patterns $\pat$ and a set of antipatterns $\apat$,
the score $s$ to judge an input $x_1, x_2$ is the difference between the maximum probability $m_{\mathrm{pos}}$
that any pattern forms a
felicitous
statement and the maximum probability $m_{\mathrm{neg}}$
that any antipattern forms a
felicitous
statement:
\begin{align*}
	m_{\mathrm{pos}} = \max_{\p\in\pat} P_{\textsc{fel}}(z = 1 \mid \p(x_1, x_2)) \\
	m_{\mathrm{neg}} = \max_{\anti\in\apat} P_{\textsc{fel}}(z = 1 \mid \anti(x_1, x_2)) \\
	s(x_1, x_2) = m_{\mathrm{pos}} - m_{\mathrm{neg}}
\end{align*}
As in \textsc{nli}, the final decision uses a threshold $\thr$:
\begin{equation*}
	D_{\textsc{pat}}^\thr(x_1, x_2) = \begin{cases}
		1, & \text{if $s(x_1, x_2) > \thr$}\\
		0, & \text{otherwise}
	\end{cases}
\end{equation*}
This corresponds to requiring that $m_{\mathrm{pos}}$ be higher than $m_{\mathrm{neg}}$ by a margin $\thr$, i.e.,
$D_{\textsc{pat}}^\thr(x_1, x_2) = 1$ iff $m_{\mathrm{pos}} > m_{\mathrm{neg}} + \thr$.

As \citet{bouraoui20} did not use antipatterns,
they defined $m_{\mathrm{neg}}$ as the maximum probability
for any pattern to form an
infelicitous
statement.
To estimate the usefulness of antipatterns,
we evaluate both possibilities,
marking systems that use both patterns and antipatterns with ${\pat\apat}$
and those that only use patterns with ${\pat}$.

The use of a threshold is another novel component,
i.e., \citet{bouraoui20} virtually set $\thr=0$.
We discuss the influence of $\thr$ in \cref{sec:results}.

We train all pattern-based approaches by minimizing the negative log-likelihood $\mathcal{L}_{\textsc{pat}}$
that patterns $\Phi$ produce
felicitous
statements for valid entailments ($y=1$) and
infelicitous
statements for invalid entailments ($y=0$) from the training data $\mathcal{T}$,
and vice versa for antipatterns $\Psi$:
\begin{align*}
	\begin{split}
		&\mathcal{L}_{\textsc{pat}}(\mathcal{T}, \pat, \apat) =\\
		&\quad \sum_{(x_1, x_2, y)\in\mathcal{T}} \mathcal{L}_{\pat}(x_1, x_2, y) + \mathcal{L}_{\apat}(x_1, x_2, 1-y)
	\end{split}
\end{align*}
with
\begin{align*}
	\begin{split}
		&\mathcal{L}_{\Omega}(x_1, x_2, y) =\\
		&\quad -\frac{1}{\abs{\Omega}}\sum_{\omega\in \Omega}\log(P_{\textsc{fel}}(z = y \mid \omega(x_1, x_2)))
	\end{split}
\end{align*}

\section{Experiments}
\label{sec:experiments}

\begin{table}
	\centering
	\begin{tabular}{llrr}
		\toprule
		&& Levy/Holt & SherLIiC \\
		\midrule
		\multicolumn{1}{l|}{\multirow{2}{*}{\dev{1}}}&train & 4,388 & 797 \\
		\multicolumn{1}{l|}{}&\dev{2}  & 1,098 & 201 \\
		\midrule
		\multicolumn{2}{c}{test} & 12,921 & 2,990 \\
		\bottomrule
	\end{tabular}
	\caption{Data split sizes as used in our experiments.}
	\label{tab:data}
\end{table}

We evaluate on two benchmarks:
\begin{enumerate*}[label=(\roman*)]
	\item Levy/Holt's dataset \citep{levy-dagan-2016-annotating,holt18} and
	\item SherLIiC \citep{schmitt-schutze-2019-sherliic}.
\end{enumerate*}
For both filtering and classification, we employ RoBERTa-base \citep{liu19}.
For classification only, we also report results for
RoBERTa-large.

\begin{figure*}[ht]
	\centering
	\begin{subfigure}[b]{\linewidth}
		\centering
		\includegraphics[height=14em]{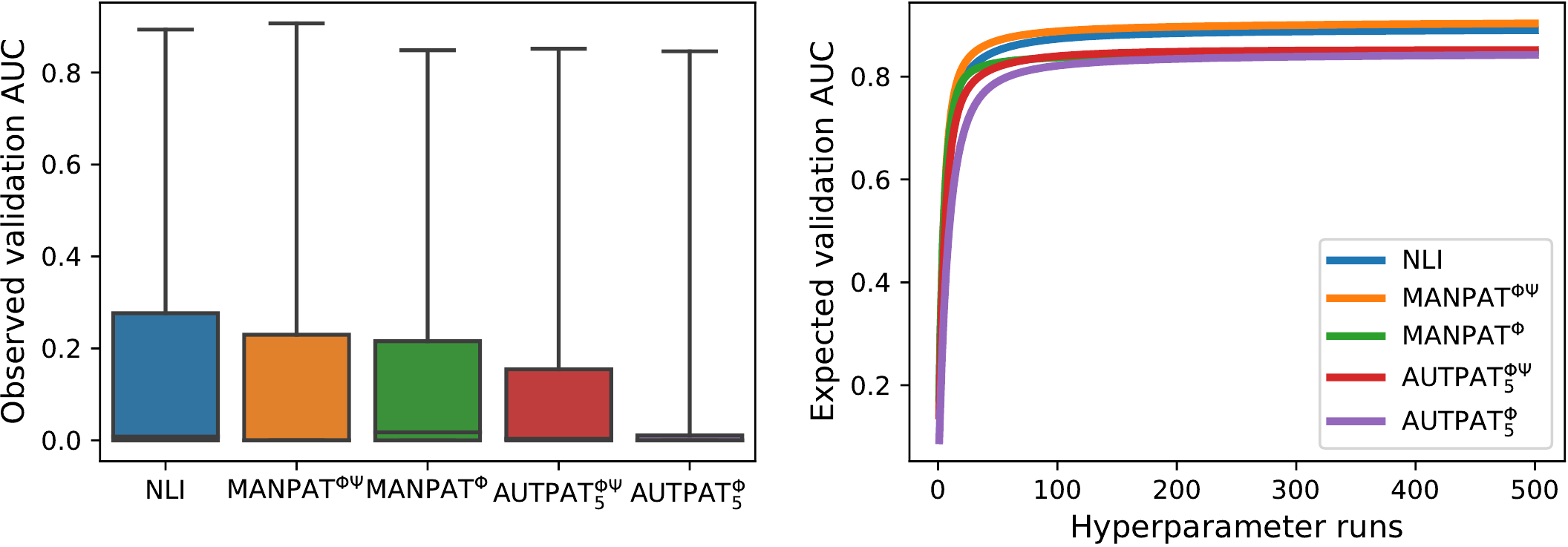}
		\caption{Levy/Holt \dev{2} RoBERTa-base}
		\label{fig:val_perf_levy_holt}
	\end{subfigure}
	
	\begin{subfigure}[b]{\linewidth}
		\centering		\includegraphics[height=14em]{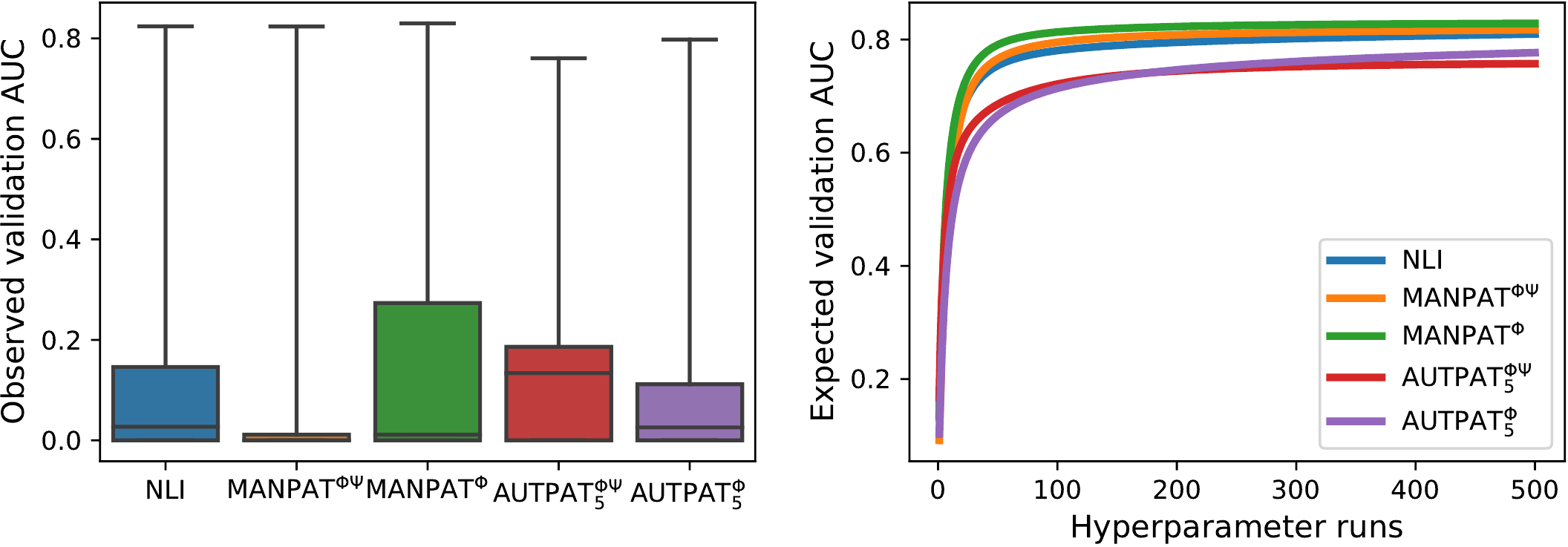}
		\caption{SherLIiC \dev{2} RoBERTa-base}
		\label{fig:val_perf_sherliic}
	\end{subfigure}
	\caption{Validation performance distribution of different datasets across different hyperparameter runs (left) and expected validation performance per number of tested hyperparameter configurations as proposed by \citet{dodge-etal-2019-show} (right). Performance is measured as the area under the precision-recall curve for precision values${}\geq 0.5$. The Boxes represent 75\%{} of the respective data points; a black line indicates the median, whiskers extend to the maximum value.}
	\label{fig:violin_val_perf}
\end{figure*}

\subsection{Data processing}

For both datasets, previous work has established a dev/test split.
For Levy/Holt, it was defined in \citep{hosseini-etal-2018-learning};
for SherLIiC, we use the original one from \citep{schmitt-schutze-2019-sherliic}.
For comparison with previous work, we keep the test portion as is
and split the dev portion further into 80\%{} for training and 20\%{} for development.
We call the new, smaller dev sets \dev{2} and the original dev sets \dev{1}.
See \cref{tab:data} for data split sizes.

\para{Levy/Holt.}
An instance in Levy/Holt has two sentences,
each consisting of two shared noun phrases (the arguments)
and a verbal expression, in which the two sentences differ.
As the verbal expressions can contain auxiliaries or negation, they often consist of multiple tokens.
Originally, one argument is replaced with a WordNet \citep{miller95} type in one of the sentences
to make the entailment more general during annotation,
but we use a version of the dataset provided by \citet{hosseini-etal-2019-duality}
where both sentences have concretely instantiated arguments.
For example, consider \cref{tab:error_analysis} (c).
\emph{Athena} was masked as the WordNet synset \emph{deity} during benchmark annotation
but we use the original sentences as shown in \cref{tab:error_analysis}
for all classifiers without further modification.

For the automatic pattern search in \textsc{autpat},
we look for sentences that mention verbatim the two verbal expressions of any instance from \dev{1}.
For the ranking, we take the last token of a verbal expression as representative for the whole.
This has the advantage that we can query the LM with a single $\spectok{mask}$ token
and compare a single token to the $k=100$ most probable predictions.
We take the last token because it usually is the main verb.

\para{SherLIiC.}
For classification, we
use SherLIiC's automatically generated sentences
that were used for annotation during benchmark creation.
The arguments in SherLIiC are entity types from
Freebase \citep{bollacker08}.
As such, they can be replaced by any Freebase entity with matching type.
For example, consider \cref{tab:error_analysis} (a);
the arguments \emph{Germany} and \emph{Côte d'Ivoire} were originally masked as \emph{location[A]} and \emph{location[B]} during annotation,
but annotators also saw three randomly chosen instantiations for both \emph{A} (\emph{Germany / Syria / USA}) and \emph{B} (\emph{Côte d'Ivoire / UK / Italy}) for context.
From the three examples provided in SherLIiC for each argument,
we choose the first one to form sentences with concretely instantiated arguments.

For the automatic pattern search in \textsc{autpat},
we make use of the greater flexibility offered by the lemmatized representations in SherLIiC.
As we are interested in statements that can be made in any way in a text,
we search for sentences
that mention the two predicates of a SherLIiC \dev{1} instance in any inflected form.
For the ranking, we again consider the predicate representative for the whole verbal expression.
We thus use the predicate lemma and otherwise proceed as described above.

\subsection{Training details}

We train all our classifiers for 5 epochs with  Adam \citep{kingma15}
and a mini-batch size of 10
(resp.\ 2)
for RoBERTa-base (resp. -large).
We randomly sample 500 configurations for the remaining hyperparameters (see \cref{app:hparam}).
For a fair comparison,
we evaluate all our approaches with the same configurations.

\begin{table}[t]
	\centering
	\small
	\begin{tabular}{l@{\, }lrrrr}
		\toprule
		&& \textsc{auc} & P & R & F1 \\
		\midrule
		\multicolumn{6}{l}{{\tiny baselines}}\\
		\multicolumn{2}{l}{\citet{hosseini-etal-2018-learning}} & 16.5 & -- & -- & -- \\
		\multicolumn{2}{l}{\citet{hosseini-etal-2019-duality}} & \textbf{18.7} & -- & -- & -- \\
		\midrule
		\multicolumn{6}{l}{{\tiny RoBERTa-base}}\\
		\textsc{nli} & {\tiny ($\thr=\phantom{-}0.0052$)} & 72.6 & 68.7 & \textbf{75.3} & 71.9 \\
		$\textsc{manpat}^{\pat\apat}$ & {\tiny ($\thr = -0.0909$)} & \textbf{76.9} & \textbf{78.7} & 66.4 & \textbf{72.0} \\
		$\textsc{manpat}^{\pat}$ & {\tiny ($\thr = \phantom{-}0.5793$)} & 71.2 & 74.4 & 61.2 & 67.1 \\
		$\textsc{autpat}_{5}^{\pat\apat}$ & {\tiny ($\thr=-0.1428$)} & 63.7 & 71.0 & 58.8 & 64.3 \\
		$\textsc{autpat}_{5}^{\pat}$ & {\tiny ($\thr = -0.0592$)} & 65.4 & 68.0 & 63.3 & 65.5 \\
		\midrule
		\multicolumn{6}{l}{{\tiny RoBERTa-large}}\\
		\textsc{nli} & {\tiny ($\thr=\phantom{-}0.0016$)} & 75.5 & 73.5 & 73.7 & 73.6\\
		$\textsc{manpat}^{\pat\apat}$ & {\tiny ($\thr=\phantom{-}0.1156$)} & \textbf{83.9} & \textbf{84.8} & 70.1 & \textbf{76.7} \\
		$\textsc{manpat}^{\pat}$ & {\tiny ($\thr=-0.8457$)} & 77.8 & 67.9 & \textbf{81.5} & 74.1 \\
		$\textsc{autpat}_{5}^{\pat\apat}$ & {\tiny ($\thr=-0.0021$)} & 70.4 & 75.7 & 60.7 & 67.4 \\
		$\textsc{autpat}_{5}^{\pat}$ & {\tiny ($\thr=-0.9197$)} & 66.5 & 61.8 & 74.4 & 67.5 \\
		\bottomrule
	\end{tabular}
	\caption{Levy/Holt test.
		\textsc{auc} denotes the area under the precision-recall curve for precision${}\geq0.5$. All results in \%{}. Bold means best result per column and block.}
	\label{tab:levyholt}
\end{table}

\begin{table}[t]
	\centering
	\small
	\begin{tabular}{l@{\, }lrrrr}
		\toprule
		&& \textsc{auc} & P & R & F\textsubscript{1} \\
		\midrule
		\multicolumn{6}{l}{\tiny baselines}\\
		\multicolumn{2}{l}{Lemma} & -- & \textbf{90.7} & 8.9 & 16.1 \\
		\multicolumn{2}{l}{w2v+untyped\_{}rel} & -- & 52.8 & 69.5 & 60.0 \\
		\multicolumn{2}{l}{w2v+tsg\_rel\_emb} & -- & 51.8 & \textbf{72.7} & \textbf{60.5} \\
		\midrule
		\multicolumn{6}{l}{\tiny RoBERTa-base}\\
		\textsc{nli} & {\tiny ($\thr = \phantom{-}0.3878$)} & 65.8 & \textbf{67.0} & 66.1 & 66.5 \\
		$\textsc{manpat}^{\pat\apat}$ & {\tiny ($\thr = -0.3324$)} & 66.4 & 60.9 & 78.8 & 68.7 \\
		$\textsc{manpat}^\pat$ & {\tiny ($\thr = -0.4812$)} & \textbf{69.2} & 62.0 & \textbf{81.2} & \textbf{70.3} \\
		$\textsc{autpat}_{5}^{\pat\apat}$ & {\tiny ($\thr = -0.4694$)} & 67.4 & 61.8 & 75.6 & 68.0 \\
		$\textsc{autpat}_{5}^{\pat}$ & {\tiny ($\thr = -0.7042$)} & 67.3 & 56.6 & 82.6 & 67.2 \\
		\midrule
		$\textsc{autcur}_{5}^{\pat}$ & {\tiny ($\thr = -0.7524$)} & \textbf{69.5} & 56.3 & \textbf{89.6} & \textbf{69.2} \\
		$\textsc{autarg}_{5}^{\pat}$ & {\tiny ($\thr = -0.7461$)} & 65.2 & \textbf{61.9} & 75.6 & 68.1 \\
		\midrule
		\multicolumn{6}{l}{\tiny RoBERTa-large}\\
		\textsc{nli} & {\tiny ($\thr = \phantom{-}0.0025$)} & 68.3 & 60.5 & \textbf{85.5} & 70.9 \\
		$\textsc{manpat}^{\pat\apat}$ & {\tiny ($\thr = -0.0956$)} & \textbf{74.4} & \textbf{66.0} & 80.8 & \textbf{72.6} \\
		$\textsc{manpat}^{\pat}$ & {\tiny ($\thr = -0.6641$)} & 64.6 & 58.1 & 79.0 & 67.0 \\
		$\textsc{autpat}_{5}^{\pat\apat}$ & {\tiny ($\thr = -0.9889$)} & 68.6 & 61.9 & 75.5 & 68.0 \\
		$\textsc{autpat}_{5}^{\pat}$ & {\tiny ($\thr = -0.5355$)} &56.8 & 61.5 & 66.1 & 63.7 \\
		\bottomrule
	\end{tabular}
	\caption{SherLIiC test. Baseline results from
          \citep{schmitt-schutze-2019-sherliic}. Table
          format: see \cref{tab:levyholt}.}
	\label{tab:sherliic}
\end{table}

\begin{table*}[t]
	\centering
	\small
	\begin{tabularx}{\linewidth}{cXll}
		\toprule
		\multicolumn{2}{l}{\textit{Automatically retrieved patterns (with SherLIiC \dev{1})}} & {prem} & {hypo} \\
		\midrule
		rank 1 & In North America, where the "atypical" forms of community-\textbf{hypo} pneumonia are & acquired & acquired \\
		& becoming more common, macrolides (such as azithromycin), and doxycycline have  displaced amoxicillin as first-line outpatient treatment for community-\textbf{prem} pneumonia. && \\
		rank 5 & This area now consists of \ldots{} the Yukon  Territory (\textbf{prem} 1898) \ldots{} and Nunavut & created & created in \\
		& (\textbf{hypo} 1999). & & \\
		rank 12 & For example, \ldots{} 訪問 "\textbf{prem}" is composed of 訪 "to visit" and 問 "to \textbf{hypo}". & interview & ask \\
		\midrule
		\multicolumn{2}{l}{\textit{Handcrafted patterns}} \\
		\midrule
		(a) & \textsc{pargl} \textbf{prem} \textsc{pargr}, which means that \textsc{hargl} \textbf{hypo} \textsc{hargr}. \\
		(b) & It is not the case that \textsc{hargl} \textbf{hypo} \textsc{hargr}, let alone that \textsc{pargl} \textbf{prem} \textsc{pargr}. \\
		(c) & \textsc{hargl} \textbf{hypo} \textsc{hargr} because \textsc{pargl} \textbf{prem} \textsc{pargr}. \\
		(d) & \textsc{pargl} \textbf{prem-negated} \textsc{pargr} because \textsc{hargl} \textbf{hypo-negated} \textsc{hargr}. \\
		(e) & \textsc{hargl} \textbf{hypo-negated} \textsc{hargr}, which means that \textsc{pargl} \textbf{prem-negated} \textsc{pargr}. \\
		\bottomrule
	\end{tabularx}
	\caption{Examples of automatically retrieved and
          ranked
 \textsc{autpat}
          patterns (top) and  handcrafted \textsc{manpat}
          patterns   (bottom). {prem}/{hypo} = original fillers as found in the corpus.
          \textsc{pargl}/\textsc{hargl} = placeholder for left argument of premise/hypothesis;
          \textsc{pargr}/\textsc{hargr} = right argument.}
	\label{tab:pattern_comparison}
\end{table*}

\section{Results and Discussion}
\label{sec:results}

\subsection{Hyperparameter robustness}

Following previous work \citep{hosseini-etal-2018-learning,hosseini-etal-2019-duality},
we use the area under the precision-recall curve (\textsc{auc}) restricted to precision values${}\geq 0.5$
as criterion for model selection.

\cref{fig:violin_val_perf} (left) shows the distribution of \dev{2} performance for  500 randomly sampled runs with RoBERTa-base.
Most hyperparameters perform poorly,
suggesting that hyperparameter search is crucial.
For Levy/Holt, \textsc{nli} is  strong whereas for SherLIiC
handcrafted $\textsc{manpat}^{\pat}$ patterns   have a clearer advantage.
For SherLIiC, 
the combination of automatically generated patterns and antipatterns $\textsc{autpat}_{5}^{\pat\apat}$
exhibits the highest median performance and the second-highest upper quartile,
making it together with $\textsc{manpat}^{\pat}$
the most robust to different hyperparameters,
although its top performance is lower compared to the others.
For all methods, only very few hyperparameter sets achieve top performances.
For both datasets, however,
a well-performing configuration is found after
fewer than 100 sampled runs (\cref{fig:violin_val_perf}, right).
Considering that $\textsc{autpat}$ requires an LM to rank thousands of patterns,
these results suggest that, for LIiC, available GPU hours should be spent on automatic hyperparameter rather than pattern search.
With its manually written patterns,
\textsc{manpat} does not need additional GPU hours for pattern search
and still, on average, performs better.

\subsection{Best hyperparameter configurations}

For the best found configuration for each method,
we not only report \textsc{auc},
which provides a general picture of a scoring method's precision-recall trade-off,
but also the concrete precision, recall, and F1 for the actual classification
after applying a threshold $\thr$.
For this we tune $\thr$ on \dev{2}  for optimal F1.
\Cref{tab:levyholt,tab:sherliic} show the results.

On both datasets,  our methods outperform all previous work (sometimes by a large margin),
thus establishing a new state of the art.
For SherLIiC+RoBERTa-base, the strong but simple \textsc{nli} system is consistently outperformed
by all pattern-based approaches,
showing that well-chosen patterns and antipatterns can be helpful for LIiC.
For SherLIiC+RoBERTa-large and also generally on Levy/Holt's dataset, \textsc{nli} is more competitive,
but the combination of handcrafted patterns and antipatterns $\textsc{manpat}^{\pat\apat}$
still performs better in these cases.

The use of antipatterns does not consistently lead to better performance
for all combinations of dataset, LM variant (base vs.\ large), and pattern set (\textsc{manpat} vs.\ \textsc{autpat}).
They do, however, consistently bring gains for some combinations,
e.g., \textsc{manpat} on Levy/Holt and \textsc{autpat} on SherLIiC.
Moreover, antipatterns are essential for achieving top performance,
i.e., the new state of the art, on both datasets.

Most of  the threshold values $\thr$ (tuned on \dev{2})
are far from their traditional values, $0.5$ for \textsc{nli} and $0.0$ for patterns.
\textsc{nli} classifiers' probability estimates are often too confident,
resulting in values close to $0$ and $1$.
To ``correct'' cases where a very small value is assigned to a valid entailment,
optimal thresholds are often close to $0$ instead of $0.5$.
Analogously, most pattern-based approaches opt for a negative $\thr$,
which means that instead of requiring a margin between
$m_{\mathrm{pos}}$ and $m_{\mathrm{neg}}$
(boosting precision), they 
make more positive predictions and boost recall.
Low recall is a key problem in LIiC (cf.\ \citet{levy-dagan-2016-annotating}).
Tuning a threshold increases the models' flexibility in this aspect.

\begin{table}[t]
	\centering
	\small
	\begin{tabular}{lrrrr}
		\toprule
		&\multicolumn{2}{c}{$\pat\apat$}&\multicolumn{2}{c}{$\pat$}\\
		\cmidrule(lr){2-3}\cmidrule(lr){4-5}
		$n$ & \textsc{auc} & F1 & \textsc{auc} & F1\\
		\midrule
		5 & 67.4 & 68.0 & 67.3 & 67.2 \\
		15 & \textbf{70.0} & \textbf{68.7} & \textbf{73.1} & \textbf{69.4} \\
		25 & 63.5 & 67.3 & 69.0 & 68.7 \\
		50 & 66.3 & 65.6 & 67.4 & 67.6 \\
		\bottomrule
	\end{tabular}
	\caption{RoBERTa-base+$\textsc{autpat}_{n}$ results on SherLIiC test for different $n$ values. Hyperparameters were tuned for the corresponding $\textsc{autpat}_{5}$ method on \dev{2}.}
	\label{tab:autpat_n_sherliic}
\end{table}

\section{Analysis}
\subsection{Number of patterns}
\label{sec:autpat_k}

\cref{sec:results} shows that automatic  patterns do not
beat handcrafted patterns for LIiC.
However, automatic patterns have one
major advantage:
in contrast to manual patterns, their number can be
easily increased.
We therefore investigate the impact of the 
hyperparameter $n$ for $\textsc{autpat}_{n}$.

\cref{tab:autpat_n_sherliic} shows that  too many
patterns is as bad as too few.  $\textsc{autpat}_{15}$
is the sweetspot:
on SherLIiC, it outperforms
all other RoBERTa-base methods on 
\textsc{auc} and closely approaches  the
otherwise best method $\textsc{manpat}^{\pat}$ on F1.

\subsection{Pattern analysis}
\label{sec:pattern_analysis}

Handcrafted patterns mostly outperform automatic ones (\cref{sec:results}).
A larger number $n$ of patterns only has a small effect (\cref{sec:autpat_k}).
We therefore take a closer look at  automatic  and
manual patterns.
\cref{tab:pattern_comparison} shows 
all handcrafted and
a sample of highly ranked automatic patterns.

It is striking how specific the automatically retrieved contexts are;
especially for the highest ranks (exemplified by ranks 1 and 5) only a narrow set of verbs seems plausible from a human perspective.
It is only at rank 12 that we find a more general context
and it
arguably even displays some semantic reasoning.
There certainly are verbs
that are not compatible with the meaning of \textit{visit},
but this context allows for a wide range of plausible verbs
and even mentions composition of meaning.

The handcrafted patterns, in contrast,
all capture some general aspect of entailment,
which might be the reason they generalize better.
Moreover, they also have placeholder slots for the verb arguments,
which could be an advantage as these represent a verb's original context.
Only accepting corpus sentences in which
the verbs occur with the same arguments
as in the  dataset
is too restrictive.

We therefore conduct the following experiment:
We manually go through the 100 highest-ranked automatically created patterns and identify 5 contexts that could accommodate arguments without changing the overall sentence structure.
We also try to pick patterns that are different enough from each other to avoid redundancy.
As a baseline,
the method $\textsc{autcur}_{5}^{\pat}$ uses these manually curated patterns as is.
We then rewrite the patterns such that they include placeholders for verb arguments,
e.g.,
``\textit{The original aim of de Garis' work was to \textbf{prem} the field of "brain building" (a term of his invention) and to "\textbf{hypo} a trillion dollar industry within 20 years".}''
becomes
``\textit{The original aim of their work was that "\textsc{pargl} \textbf{prem} \textsc{pargr}" and that "\textsc{hargl} \textbf{hypo} \textsc{hargr} within 20 years".}''
with \textsc{pargl} / \textsc{pargr} (\textsc{hargl} / \textsc{hargr})
the placeholder for the left / right argument of the premise (hypothesis).
See \cref{tab:pattern_rewriting} in the appendix for the complete list.
$\textsc{autarg}_{5}^{\pat}$ is based on these rewritten patterns.
We try the same 500 hyperparameter configurations as for the other RoBERTa-base approaches
and include results for the best configuration (chosen on \dev{2}) in \cref{tab:sherliic}.
We find that manually curating automatically ranked patterns  helps performance.
$\textsc{autcur}_{5}^{\pat}$ outperforms $\textsc{autpat}_{5}^{\pat}$ on \textsc{auc} and F1,
reducing the gap to handcrafted patterns (i.e., $\textsc{manpat}^{\pat}$).
This is probably due to the variety we enforced when handpicking the patterns.

Surprisingly, adding arguments decreases performance.
Possibly, our modifications make the patterns less
fluent
or the arguments
that are filled into the placeholders during training and evaluation
do not fit well into the contexts,
which still are rather specific.

\begin{table}[t]
	\centering
	\small
	\begin{tabularx}{\linewidth}{lX}
		\toprule
		(a)	& \textit{Germany is occupying Côte d'Ivoire} \\
		& $\Rightarrow$ \textit{Germany is remaining in Côte d'Ivoire} \\
		Sh	& \predictions{1}{\wrong{0}}{\correct{1}}{\wrong{0}}{\correct{1}}{\correct{1}} \\
		\midrule
		(b)	& \textit{Ford awarded him the medal} \\
		& $\Rightarrow$ \textit{Ford was awarded a medal} \\
		L/H	& \predictions{0}{\wrong{1}}{\correct{0}}{\correct{0}}{\wrong{1}}{\wrong{1}} \\
		\midrule
		(c)	& \textit{Athena was worshiped in Athens} \\
		& $\Rightarrow$ \textit{Athena was the goddess of Athens} \\
		L/H	& \predictions{0}{\correct{0}}{\correct{0}}{\wrong{1}}{\wrong{1}}{\wrong{1}} \\
		\midrule
		(d)	& \textit{Pyrrhus was beaten by the romans} \\
		& $\Rightarrow$ \textit{Pyrrhus fought the romans} \\
		L/H	& \predictions{1}{\correct{1}}{\wrong{0}}{\wrong{0}}{\wrong{0}}{\correct{1}} \\
		\midrule\midrule
		(e)	& \textit{England national rugby union team is playing against Denver Broncos} \\
		& $\Rightarrow$ \textit{England national rugby union team is beating Denver Broncos} \\
		Sh	& \predictions{0}{\wrong{1}}{\wrong{1}}{\wrong{1}}{\wrong{1}}{\wrong{1}} \\
		\midrule
		(f)	& \textit{Polk negotiated with Britain} \\
		& $\Rightarrow$ \textit{Polk made peace with Britain} \\
		L/H	& \predictions{0}{\wrong{1}}{\wrong{1}}{\wrong{1}}{\wrong{1}}{\wrong{1}} \\
		\bottomrule
	\end{tabularx}
	\caption{Qualitative error analysis of the RoBERTa-base models from \cref{tab:levyholt,tab:sherliic} on the \dev{2} split of SherLIiC (Sh) and Levy/Holt (L/H). Pattern-based predictions are listed in the format $\pat\apat$ / $\pat$. Correct predictions are green; errors are underlined and red.}
	\label{tab:error_analysis}
\end{table}

\subsection{Error analysis}
\label{sec:error_analysis}

\Cref{tab:error_analysis} displays a selection of the \dev{2} sets of our two benchmarks
along with the predictions of all our approaches.

The first four examples indicate how \textsc{nli} differs from pattern approaches.
Example (a) involves the common sense knowledge that occupying a territory implies remaining there.
This might be learned from patterns more easily
as these patterns might resemble contexts -- seen during pretraining --
that describe how long a military force remained during an occupation.
Putting the inference candidate (b) into a pattern-generated context avoids
being fooled by the high similarity of the two sentences.
Only the handcrafted patterns can make sense of the important details in this construction.

In contrast, (c) and (d) are difficult for our pattern approaches
whereas \textsc{nli} gets them right.
We hypothesize that
the problem stems from linking the two sentences into one.
An entailment pattern ideally represents a derivation of the hypothesis from the premise.
One may wrongly conclude that (c) \textit{Athena was the goddess of Athens only because she was worshiped there},
by neglecting the possibility that there are others that are equally worshiped.
In the same way, (d) is unlikely to be found in an argumentative text.
While it is clear that there can be no beating without a fight,
one would hardly argue that \textit{Pyrrhus fought the romans because they beat him}.
This particular reasoning calls for additional explanations like
\textit{Pyrrhus must have fought the romans because I know that they beat him}.
This analysis serves as inspiration for further improvements of entailment patterns.

The last two examples (e) and (f) are difficult for all approaches.
It seems to be a particular challenge
to identify open situations like a sports match or a negotiation
where multiple outcomes are possible
and distinguish them from cases where one particular outcome is inevitable.

\section{Conclusion}

We proposed and evaluated three approaches to the task of lexical inference in context (LIiC)
based on pretrained language models (LMs).
In particular, we found that putting an inference candidate into a pattern-generated context
mostly increases performance compared to a standard sequence classification approach.
Concrete performance, however, also depends on the particular dataset, used LM (variant), and pattern set.
We introduced the concept of antipatterns, which express the negative class of a binary classification,
and found that they often lead to performance gains for LIiC.
We set a new state of the art for LIiC
and conducted an extensive analysis of our approaches.
Notably, we found that
automatically created patterns can perform nearly as well as handcrafted ones
if we either use the right number $n$ of patterns
or manually identify the right subset of them.
Promising directions for future work are the investigation of alternative automatic pattern generation methods
or a better modeling of the remaining challenges we described in our error analysis (\cref{sec:error_analysis}).

\section*{Acknowledgments}
We gratefully acknowledge a Ph.D. scholarship
awarded to the first author by the German Academic Scholarship Foundation (Studienstiftung des
deutschen Volkes). This work was supported by the
BMBF as part of the project MLWin (01IS18050).

\bibliography{anthology,references}

\begin{thebibliography}{47}
\expandafter\ifx\csname natexlab\endcsname\relax\def\natexlab#1{#1}\fi

\bibitem[{Amrami and Goldberg(2018)}]{amrami-goldberg-2018-word}
Asaf Amrami and Yoav Goldberg. 2018.
\newblock \href {https://doi.org/10.18653/v1/D18-1523} {Word sense induction
  with neural bi{LM} and symmetric patterns}.
\newblock In \emph{Proceedings of the 2018 Conference on Empirical Methods in
  Natural Language Processing}, pages 4860--4867, Brussels, Belgium.
  Association for Computational Linguistics.

\bibitem[{Berant et~al.(2011)Berant, Dagan, and
  Goldberger}]{berant-etal-2011-global}
Jonathan Berant, Ido Dagan, and Jacob Goldberger. 2011.
\newblock \href {https://www.aclweb.org/anthology/P11-1062} {Global learning of
  typed entailment rules}.
\newblock In \emph{Proceedings of the 49th Annual Meeting of the Association
  for Computational Linguistics: Human Language Technologies}, pages 610--619,
  Portland, Oregon, USA. Association for Computational Linguistics.

\bibitem[{Bollacker et~al.(2008)Bollacker, Evans, Paritosh, Sturge, and
  Taylor}]{bollacker08}
Kurt Bollacker, Colin Evans, Praveen Paritosh, Tim Sturge, and Jamie Taylor.
  2008.
\newblock \href {https://doi.org/10.1145/1376616.1376746} {Freebase: A
  collaboratively created graph database for structuring human knowledge}.
\newblock In \emph{Proceedings of the 2008 ACM SIGMOD International Conference
  on Management of Data}, SIGMOD '08, page 1247–1250, New York, NY, USA.
  Association for Computing Machinery.

\bibitem[{Bouraoui et~al.(2020)Bouraoui, Camacho-Collados, and
  Schockaert}]{bouraoui20}
Zied Bouraoui, Jose Camacho-Collados, and Steven Schockaert. 2020.
\newblock Inducing relational knowledge from bert.
\newblock In \emph{Thirty-Fourth AAAI Conference on Artificial Intelligence
  (AAAI)}.

\bibitem[{Brown et~al.(2020)Brown, Mann, Ryder, Subbiah, Kaplan, Dhariwal,
  Neelakantan, Shyam, Sastry, Askell, Agarwal, Herbert-Voss, Krueger, Henighan,
  Child, Ramesh, Ziegler, Wu, Winter, Hesse, Chen, Sigler, Litwin, Gray, Chess,
  Clark, Berner, McCandlish, Radford, Sutskever, and Amodei}]{brown20}
Tom~B. Brown, Benjamin Mann, Nick Ryder, Melanie Subbiah, Jared Kaplan,
  Prafulla Dhariwal, Arvind Neelakantan, Pranav Shyam, Girish Sastry, Amanda
  Askell, Sandhini Agarwal, Ariel Herbert-Voss, Gretchen Krueger, Tom Henighan,
  Rewon Child, Aditya Ramesh, Daniel~M. Ziegler, Jeffrey Wu, Clemens Winter,
  Christopher Hesse, Mark Chen, Eric Sigler, Mateusz Litwin, Scott Gray,
  Benjamin Chess, Jack Clark, Christopher Berner, Sam McCandlish, Alec Radford,
  Ilya Sutskever, and Dario Amodei. 2020.
\newblock \href {https://arxiv.org/abs/2005.14165} {Language models are
  few-shot learners}.
\newblock \emph{Computing Research Repository}, arXiv:2005.14165.

\bibitem[{Chklovski and Pantel(2004)}]{chklovski-pantel-2004-verbocean}
Timothy Chklovski and Patrick Pantel. 2004.
\newblock \href {https://www.aclweb.org/anthology/W04-3205} {{V}erb{O}cean:
  Mining the web for fine-grained semantic verb relations}.
\newblock In \emph{Proceedings of the 2004 Conference on Empirical Methods in
  Natural Language Processing}, pages 33--40, Barcelona, Spain. Association for
  Computational Linguistics.

\bibitem[{Dagan et~al.(2013)Dagan, Roth, Sammons, and Zanzotto}]{dagan13}
Ido Dagan, Dan Roth, Mark Sammons, and Fabio~Massimo Zanzotto. 2013.
\newblock \emph{Recognizing textual entailment: Models and applications}.
\newblock Morgan \& Claypool Publishers.

\bibitem[{Devlin et~al.(2019)Devlin, Chang, Lee, and
  Toutanova}]{devlin-etal-2019-bert}
Jacob Devlin, Ming-Wei Chang, Kenton Lee, and Kristina Toutanova. 2019.
\newblock \href {https://doi.org/10.18653/v1/N19-1423} {{BERT}: Pre-training of
  deep bidirectional transformers for language understanding}.
\newblock In \emph{Proceedings of the 2019 Conference of the North {A}merican
  Chapter of the Association for Computational Linguistics: Human Language
  Technologies, Volume 1 (Long and Short Papers)}, pages 4171--4186,
  Minneapolis, Minnesota. Association for Computational Linguistics.

\bibitem[{Dodge et~al.(2019)Dodge, Gururangan, Card, Schwartz, and
  Smith}]{dodge-etal-2019-show}
Jesse Dodge, Suchin Gururangan, Dallas Card, Roy Schwartz, and Noah~A. Smith.
  2019.
\newblock \href {https://doi.org/10.18653/v1/D19-1224} {Show your work:
  Improved reporting of experimental results}.
\newblock In \emph{Proceedings of the 2019 Conference on Empirical Methods in
  Natural Language Processing and the 9th International Joint Conference on
  Natural Language Processing (EMNLP-IJCNLP)}, pages 2185--2194, Hong Kong,
  China. Association for Computational Linguistics.

\bibitem[{Fellbaum and Miller(1990)}]{fellbaum90}
Christiane Fellbaum and George~A. Miller. 1990.
\newblock \href {https://doi.org/10.1037/0033-295X.97.4.565} {Folk psychology
  or semantic entailment? comment on rips and conrad (1989)}.
\newblock \emph{Psychological Review}, 97(4):565--570.

\bibitem[{Forbes et~al.(2019)Forbes, Holtzman, and Choi}]{forbes19}
Maxwell Forbes, Ari Holtzman, and Yejin Choi. 2019.
\newblock \href {https://mindmodeling.org/cogsci2019/papers/0311/index.html}
  {Do neural language representations learn physical commonsense?}
\newblock In \emph{Proceedings of the 41th Annual Meeting of the Cognitive
  Science Society (CogSci 2019)}, pages 1753--1759.
  cognitivesciencesociety.org.

\bibitem[{Glockner et~al.(2018)Glockner, Shwartz, and
  Goldberg}]{glockner-etal-2018-breaking}
Max Glockner, Vered Shwartz, and Yoav Goldberg. 2018.
\newblock \href {https://doi.org/10.18653/v1/P18-2103} {Breaking {NLI} systems
  with sentences that require simple lexical inferences}.
\newblock In \emph{Proceedings of the 56th Annual Meeting of the Association
  for Computational Linguistics (Volume 2: Short Papers)}, pages 650--655,
  Melbourne, Australia. Association for Computational Linguistics.

\bibitem[{Hearst(1992)}]{hearst-1992-automatic}
Marti~A. Hearst. 1992.
\newblock \href {https://www.aclweb.org/anthology/C92-2082} {Automatic
  acquisition of hyponyms from large text corpora}.
\newblock In \emph{{COLING} 1992 Volume 2: The 15th {I}nternational
  {C}onference on {C}omputational {L}inguistics}.

\bibitem[{Holt(2018)}]{holt18}
Xavier~R. Holt. 2018.
\newblock Probabilistic models of relational implication.
\newblock Master's thesis, Macquarie University.

\bibitem[{Hosseini et~al.(2018)Hosseini, Chambers, Reddy, Holt, Cohen, Johnson,
  and Steedman}]{hosseini-etal-2018-learning}
Mohammad~Javad Hosseini, Nathanael Chambers, Siva Reddy, Xavier~R. Holt,
  Shay~B. Cohen, Mark Johnson, and Mark Steedman. 2018.
\newblock \href {https://doi.org/10.1162/tacl_a_00250} {Learning typed
  entailment graphs with global soft constraints}.
\newblock \emph{Transactions of the Association for Computational Linguistics},
  6:703--717.

\bibitem[{Hosseini et~al.(2019)Hosseini, Cohen, Johnson, and
  Steedman}]{hosseini-etal-2019-duality}
Mohammad~Javad Hosseini, Shay~B. Cohen, Mark Johnson, and Mark Steedman. 2019.
\newblock \href {https://doi.org/10.18653/v1/P19-1468} {Duality of link
  prediction and entailment graph induction}.
\newblock In \emph{Proceedings of the 57th Annual Meeting of the Association
  for Computational Linguistics}, pages 4736--4746, Florence, Italy.
  Association for Computational Linguistics.

\bibitem[{Kassner and Sch{\"u}tze(2020)}]{kassner-schutze-2020-negated}
Nora Kassner and Hinrich Sch{\"u}tze. 2020.
\newblock \href {https://doi.org/10.18653/v1/2020.acl-main.698} {Negated and
  misprimed probes for pretrained language models: Birds can talk, but cannot
  fly}.
\newblock In \emph{Proceedings of the 58th Annual Meeting of the Association
  for Computational Linguistics}, pages 7811--7818, Online. Association for
  Computational Linguistics.

\bibitem[{Kiela et~al.(2015)Kiela, Rimell, Vuli{\'c}, and
  Clark}]{kiela-etal-2015-exploiting}
Douwe Kiela, Laura Rimell, Ivan Vuli{\'c}, and Stephen Clark. 2015.
\newblock \href {https://doi.org/10.3115/v1/P15-2020} {Exploiting image
  generality for lexical entailment detection}.
\newblock In \emph{Proceedings of the 53rd Annual Meeting of the Association
  for Computational Linguistics and the 7th International Joint Conference on
  Natural Language Processing (Volume 2: Short Papers)}, pages 119--124,
  Beijing, China. Association for Computational Linguistics.

\bibitem[{Kingma and Ba(2015)}]{kingma15}
Diederik~P. Kingma and Jimmy Ba. 2015.
\newblock Adam: A method for stochastic optimization.
\newblock In \emph{Proceedings of the 3rd International Conference on Learning
  Representations (ICLR)}.

\bibitem[{Kotlerman et~al.(2010)Kotlerman, Dagan, Szpektor, and
  Zhitomirsky-Geffet}]{kotlerman10}
Lili Kotlerman, Ido Dagan, Idan Szpektor, and Maayan Zhitomirsky-Geffet. 2010.
\newblock Directional distributional similarity for lexical inference.
\newblock \emph{Natural Language Engineering}, 16(04):359--389.

\bibitem[{Levy and Dagan(2016)}]{levy-dagan-2016-annotating}
Omer Levy and Ido Dagan. 2016.
\newblock \href {https://doi.org/10.18653/v1/P16-2041} {Annotating relation
  inference in context via question answering}.
\newblock In \emph{Proceedings of the 54th Annual Meeting of the Association
  for Computational Linguistics (Volume 2: Short Papers)}, pages 249--255,
  Berlin, Germany. Association for Computational Linguistics.

\bibitem[{Levy and Goldberg(2014)}]{levy-goldberg-2014-dependency}
Omer Levy and Yoav Goldberg. 2014.
\newblock \href {https://doi.org/10.3115/v1/P14-2050} {Dependency-based word
  embeddings}.
\newblock In \emph{Proceedings of the 52nd Annual Meeting of the Association
  for Computational Linguistics (Volume 2: Short Papers)}, pages 302--308,
  Baltimore, Maryland. Association for Computational Linguistics.

\bibitem[{Lewis and Steedman(2013)}]{lewis-steedman-2013-combined}
Mike Lewis and Mark Steedman. 2013.
\newblock \href {https://doi.org/10.1162/tacl_a_00219} {Combined distributional
  and logical semantics}.
\newblock \emph{Transactions of the Association for Computational Linguistics},
  1:179--192.

\bibitem[{Lin(1998)}]{lin-1998-automatic-retrieval}
Dekang Lin. 1998.
\newblock \href {https://doi.org/10.3115/980691.980696} {Automatic retrieval
  and clustering of similar words}.
\newblock In \emph{36th Annual Meeting of the Association for Computational
  Linguistics and 17th International Conference on Computational Linguistics,
  Volume 2}, pages 768--774, Montreal, Quebec, Canada. Association for
  Computational Linguistics.

\bibitem[{Lin and Pantel(2001)}]{lin01}
Dekang Lin and Patrick Pantel. 2001.
\newblock \href {http://www.cs.ualberta.ca/~lindek/papers/kdd01-1.pdf} {{DIRT}:
  {D}iscovery of {I}nference {R}ules from {T}ext}.
\newblock In \emph{Proceedings of the Seventh ACM SIGKDD International
  Conference on Knowledge Discovery and Data Mining (KDD'01)}, pages 323--328,
  New York, NY, USA. ACM Press.

\bibitem[{Liu et~al.(2019)Liu, Ott, Goyal, Du, Joshi, Chen, Levy, Lewis,
  Zettlemoyer, and Stoyanov}]{liu19}
Yinhan Liu, Myle Ott, Naman Goyal, Jingfei Du, Mandar Joshi, Danqi Chen, Omer
  Levy, Mike Lewis, Luke Zettlemoyer, and Veselin Stoyanov. 2019.
\newblock \href {https://arxiv.org/abs/1907.11692} {Roberta: A robustly
  optimized bert pretraining approach}.
\newblock \emph{Computing Research Repository}, arXiv:1907.11692.

\bibitem[{Meged et~al.(2020)Meged, Caciularu, Shwartz, and Dagan}]{meged20}
Yehudit Meged, Avi Caciularu, Vered Shwartz, and Ido Dagan. 2020.
\newblock \href {https://arxiv.org/abs/2004.14979} {Paraphrasing vs
  coreferring: Two sides of the same coin}.
\newblock \emph{Computing Research Repository}, arXiv:2004.14979.

\bibitem[{Mikolov et~al.(2013)Mikolov, Chen, Corrado, and Dean}]{mikolov13}
Tomas Mikolov, Kai Chen, Greg Corrado, and Jeffrey Dean. 2013.
\newblock \href {https://arxiv.org/abs/1301.3781} {Efficient estimation of word
  representations in vector space}.
\newblock \emph{Computing Research Repository}, arXiv:1301.3781.

\bibitem[{Miller(1995)}]{miller95}
George~A. Miller. 1995.
\newblock Wordnet: A lexical database for english.
\newblock \emph{Communications of the ACM}, 38(11):39--41.

\bibitem[{Mirkin et~al.(2006)Mirkin, Dagan, and
  Geffet}]{mirkin-etal-2006-integrating}
Shachar Mirkin, Ido Dagan, and Maayan Geffet. 2006.
\newblock \href {https://www.aclweb.org/anthology/P06-2075} {Integrating
  pattern-based and distributional similarity methods for lexical entailment
  acquisition}.
\newblock In \emph{Proceedings of the {COLING}/{ACL} 2006 Main Conference
  Poster Sessions}, pages 579--586, Sydney, Australia. Association for
  Computational Linguistics.

\bibitem[{Pavlick et~al.(2015)Pavlick, Rastogi, Ganitkevitch, Van~Durme, and
  Callison-Burch}]{pavlick-etal-2015-ppdb}
Ellie Pavlick, Pushpendre Rastogi, Juri Ganitkevitch, Benjamin Van~Durme, and
  Chris Callison-Burch. 2015.
\newblock \href {https://doi.org/10.3115/v1/P15-2070} {{PPDB} 2.0: Better
  paraphrase ranking, fine-grained entailment relations, word embeddings, and
  style classification}.
\newblock In \emph{Proceedings of the 53rd Annual Meeting of the Association
  for Computational Linguistics and the 7th International Joint Conference on
  Natural Language Processing (Volume 2: Short Papers)}, pages 425--430,
  Beijing, China. Association for Computational Linguistics.

\bibitem[{Petroni et~al.(2019)Petroni, Rockt{\"a}schel, Riedel, Lewis, Bakhtin,
  Wu, and Miller}]{petroni-etal-2019-language}
Fabio Petroni, Tim Rockt{\"a}schel, Sebastian Riedel, Patrick Lewis, Anton
  Bakhtin, Yuxiang Wu, and Alexander Miller. 2019.
\newblock \href {https://doi.org/10.18653/v1/D19-1250} {Language models as
  knowledge bases?}
\newblock In \emph{Proceedings of the 2019 Conference on Empirical Methods in
  Natural Language Processing and the 9th International Joint Conference on
  Natural Language Processing (EMNLP-IJCNLP)}, pages 2463--2473, Hong Kong,
  China. Association for Computational Linguistics.

\bibitem[{Radford et~al.(2019)Radford, Wu, Child, Luan, Amodei, and
  Sutskever}]{radford19}
Alec Radford, Jeff Wu, Rewon Child, David Luan, Dario Amodei, and Ilya
  Sutskever. 2019.
\newblock Language models are unsupervised multitask learners.

\bibitem[{Roller and Erk(2016)}]{roller-erk-2016-relations}
Stephen Roller and Katrin Erk. 2016.
\newblock \href {https://doi.org/10.18653/v1/D16-1234} {Relations such as
  hypernymy: Identifying and exploiting hearst patterns in distributional
  vectors for lexical entailment}.
\newblock In \emph{Proceedings of the 2016 Conference on Empirical Methods in
  Natural Language Processing}, pages 2163--2172, Austin, Texas. Association
  for Computational Linguistics.

\bibitem[{Roller et~al.(2018)Roller, Kiela, and
  Nickel}]{roller-etal-2018-hearst}
Stephen Roller, Douwe Kiela, and Maximilian Nickel. 2018.
\newblock \href {https://doi.org/10.18653/v1/P18-2057} {Hearst patterns
  revisited: Automatic hypernym detection from large text corpora}.
\newblock In \emph{Proceedings of the 56th Annual Meeting of the Association
  for Computational Linguistics (Volume 2: Short Papers)}, pages 358--363,
  Melbourne, Australia. Association for Computational Linguistics.

\bibitem[{Schick and Sch\"{u}tze(2020)}]{schick20a}
Timo Schick and Hinrich Sch\"{u}tze. 2020.
\newblock \href {https://arxiv.org/abs/2001.07676} {Exploiting cloze questions
  for few shot text classification and natural language inference}.
\newblock \emph{Computing Research Repository}, arXiv:2001.07676.

\bibitem[{Schmitt and Sch{\"u}tze(2019)}]{schmitt-schutze-2019-sherliic}
Martin Schmitt and Hinrich Sch{\"u}tze. 2019.
\newblock \href {https://doi.org/10.18653/v1/P19-1086} {{S}her{LI}i{C}: A typed
  event-focused lexical inference benchmark for evaluating natural language
  inference}.
\newblock In \emph{Proceedings of the 57th Annual Meeting of the Association
  for Computational Linguistics}, pages 902--914, Florence, Italy. Association
  for Computational Linguistics.

\bibitem[{Schoenmackers et~al.(2010)Schoenmackers, Davis, Etzioni, and
  Weld}]{schoenmackers-etal-2010-learning}
Stefan Schoenmackers, Jesse Davis, Oren Etzioni, and Daniel Weld. 2010.
\newblock \href {https://www.aclweb.org/anthology/D10-1106} {Learning
  first-order horn clauses from web text}.
\newblock In \emph{Proceedings of the 2010 Conference on Empirical Methods in
  Natural Language Processing}, pages 1088--1098, Cambridge, MA. Association
  for Computational Linguistics.

\bibitem[{Schwartz et~al.(2015)Schwartz, Reichart, and
  Rappoport}]{schwartz-etal-2015-symmetric}
Roy Schwartz, Roi Reichart, and Ari Rappoport. 2015.
\newblock \href {https://doi.org/10.18653/v1/K15-1026} {Symmetric pattern based
  word embeddings for improved word similarity prediction}.
\newblock In \emph{Proceedings of the Nineteenth Conference on Computational
  Natural Language Learning}, pages 258--267, Beijing, China. Association for
  Computational Linguistics.

\bibitem[{Shwartz et~al.(2015)Shwartz, Levy, Dagan, and
  Goldberger}]{shwartz-etal-2015-learning}
Vered Shwartz, Omer Levy, Ido Dagan, and Jacob Goldberger. 2015.
\newblock \href {https://doi.org/10.18653/v1/K15-1018} {Learning to exploit
  structured resources for lexical inference}.
\newblock In \emph{Proceedings of the Nineteenth Conference on Computational
  Natural Language Learning}, pages 175--184, Beijing, China. Association for
  Computational Linguistics.

\bibitem[{Shwartz et~al.(2017)Shwartz, Stanovsky, and
  Dagan}]{shwartz-etal-2017-acquiring}
Vered Shwartz, Gabriel Stanovsky, and Ido Dagan. 2017.
\newblock \href {https://doi.org/10.18653/v1/S17-1019} {Acquiring predicate
  paraphrases from news tweets}.
\newblock In \emph{Proceedings of the 6th Joint Conference on Lexical and
  Computational Semantics (*{SEM} 2017)}, pages 155--160, Vancouver, Canada.
  Association for Computational Linguistics.

\bibitem[{Szpektor and Dagan(2008)}]{szpektor-dagan-2008-learning}
Idan Szpektor and Ido Dagan. 2008.
\newblock \href {https://www.aclweb.org/anthology/C08-1107} {Learning
  entailment rules for unary templates}.
\newblock In \emph{Proceedings of the 22nd International Conference on
  Computational Linguistics (Coling 2008)}, pages 849--856, Manchester, UK.
  Coling 2008 Organizing Committee.

\bibitem[{Vaswani et~al.(2017)Vaswani, Shazeer, Parmar, Uszkoreit, Jones,
  Gomez, Kaiser, and Polosukhin}]{vaswani17}
Ashish Vaswani, Noam Shazeer, Niki Parmar, Jakob Uszkoreit, Llion Jones,
  Aidan~N Gomez, \L~ukasz Kaiser, and Illia Polosukhin. 2017.
\newblock \href
  {http://papers.nips.cc/paper/7181-attention-is-all-you-need.pdf} {Attention
  is all you need}.
\newblock In I.~Guyon, U.~V. Luxburg, S.~Bengio, H.~Wallach, R.~Fergus,
  S.~Vishwanathan, and R.~Garnett, editors, \emph{Advances in Neural
  Information Processing Systems 30}, pages 5998--6008. Curran Associates, Inc.

\bibitem[{Vuli{\'c} and
  Mrk{\v{s}}i{\'c}(2018)}]{vulic-mrksic-2018-specialising}
Ivan Vuli{\'c} and Nikola Mrk{\v{s}}i{\'c}. 2018.
\newblock \href {https://doi.org/10.18653/v1/N18-1103} {Specialising word
  vectors for lexical entailment}.
\newblock In \emph{Proceedings of the 2018 Conference of the North {A}merican
  Chapter of the Association for Computational Linguistics: Human Language
  Technologies, Volume 1 (Long Papers)}, pages 1134--1145, New Orleans,
  Louisiana. Association for Computational Linguistics.

\bibitem[{Weeds and Weir(2003)}]{weeds-weir-2003-general}
Julie Weeds and David Weir. 2003.
\newblock \href {https://www.aclweb.org/anthology/W03-1011} {A general
  framework for distributional similarity}.
\newblock In \emph{Proceedings of the 2003 Conference on Empirical Methods in
  Natural Language Processing}, pages 81--88.

\bibitem[{Williams et~al.(2018)Williams, Nangia, and
  Bowman}]{williams-etal-2018-broad}
Adina Williams, Nikita Nangia, and Samuel Bowman. 2018.
\newblock \href {https://doi.org/10.18653/v1/N18-1101} {A broad-coverage
  challenge corpus for sentence understanding through inference}.
\newblock In \emph{Proceedings of the 2018 Conference of the North {A}merican
  Chapter of the Association for Computational Linguistics: Human Language
  Technologies, Volume 1 (Long Papers)}, pages 1112--1122, New Orleans,
  Louisiana. Association for Computational Linguistics.

\bibitem[{Wolf et~al.(2019)Wolf, Debut, Sanh, Chaumond, Delangue, Moi, Cistac,
  Rault, Louf, Funtowicz, Davison, Shleifer, von Platen, Ma, Jernite, Plu, Xu,
  Scao, Gugger, Drame, Lhoest, and Rush}]{huggingface}
Thomas Wolf, Lysandre Debut, Victor Sanh, Julien Chaumond, Clement Delangue,
  Anthony Moi, Pierric Cistac, Tim Rault, Rémi Louf, Morgan Funtowicz, Joe
  Davison, Sam Shleifer, Patrick von Platen, Clara Ma, Yacine Jernite, Julien
  Plu, Canwen Xu, Teven~Le Scao, Sylvain Gugger, Mariama Drame, Quentin Lhoest,
  and Alexander~M. Rush. 2019.
\newblock Huggingface's transformers: State-of-the-art natural language
  processing.
\newblock \emph{ArXiv}, abs/1910.03771.

\end{thebibliography}
\bibliographystyle{acl_natbib}

\appendix
\section{Hyperparameters}
\label{app:hparam}

We train all our classifiers for 5 epochs with the Adam optimizer \citep{kingma15}
and a mini-batch size of 10 or 2 instances for RoBERTa-base and -large, respectively.
For \textsc{autpat}\textsubscript{$n$} approaches with $n > 5$,
we distribute the available patterns and antipatterns into chunks of size 5 for training
to save memory.
During evaluation, the predictions are based on the all patterns and antipatterns.

We randomly sample 500 configurations for the remaining hyperparameters,
i.e.,
initial learning rate $\mathrm{lr}$,
weight decay $\lambda$ (L2 regularization),
and the number of batches $c$ the gradient is accumulated before each optimization step,
which virtually increases the batch size by a factor of $c$.
The hyperparameters are sampled from the following intervals:
$\mathrm{lr}\in\interval{10^{-8}}{5\cdot 10^{-2}}$, $\lambda\in\interval{10^{-5}}{10^{-1}}$,
$c\in\eset{1, 2, \dots, 10}$.
$\mathrm{lr}$ and $\lambda$ are sampled uniformly in log-space.
For a fair comparison,
we use the same 500 random configurations for all of our approaches.

As usual for Transformer models, we apply a learning rate schedule:
$\mathrm{lr}$ decreases linearly such that it reaches $0$ at the end of the last epoch.
We do not employ warm-up.

The best configurations can be seen in \cref{tab:hparam_levyholt_base,tab:hparam_levyholt_large} for Levy/Holt's dataset and \cref{tab:hparam_sherliic_base,tab:hparam_sherliic_large} for SherLIiC.

\section{Results on development sets}
\label{sec:more_results}

See \cref{tab:dev_sherliic,tab:dev_levyholt}.

\section{Varying $n$ in training and evaluation}
\label{sec:varying_n}

\begin{figure}[h!]
	\centering
	\includegraphics[width=\linewidth]{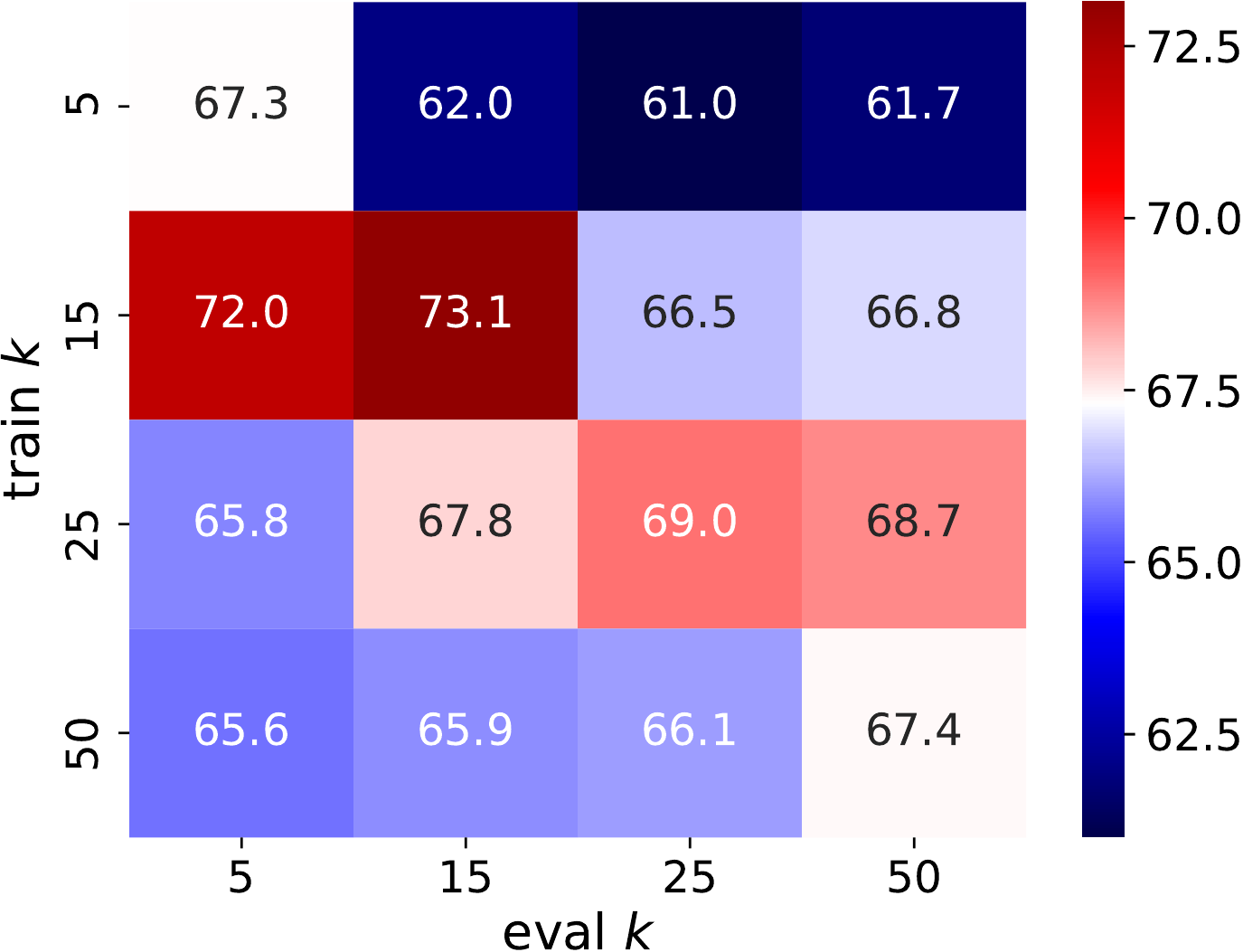}
	\caption{RoBERTa-base+$\textsc{autpat}_{k}^{\pat}$ performance on SherLIiC test for different $k$ values during training and evaluation. Same hyperparameters used for all models (as in \cref{tab:autpat_n_sherliic}). Blue marks drops, red marks gains in performance w.r.t.\ $\textsc{autpat}_{5}^{\pat}$.}
	\label{fig:autpat_n_sherliic}
\end{figure}

Another approach to make use of different values of $n$ is to vary $n$ from training to evaluation.
\Cref{fig:autpat_n_sherliic} is a visualization of the performance impact of this procedure.
The base point for the visualization (in white) is the \textsc{auc} performance of $\textsc{autpat}_{5}^{\pat}$.
We see that training with $n=50$ almost always leads to a performance drop (marked in blue) w.r.t.\ this number.
It seems generally to be catastrophic to evaluate a model with patterns that were not seen during training,
indicating that there is no generalization from seen patterns to unseen patterns even if they were chosen by the same method
and can thus be expected to be
-- at least to some extent --
similar.
In general,
this evaluation suggests that modifying $n$ after the training always leads to a drop in performance.

\section{Transfer between Datasets}

\Cref{tab:transfer} shows results on the question how well a model trained on one dataset performs on the other.
For this, we assume that the target dataset is not available at all, i.e., we do not use it at all -- neither for finding patterns in \textsc{autpat} nor for tuning the threshold $\vartheta$. We thus use the standard $\vartheta$ values, i.e., $0.5$ for \textsc{nli} and $0.0$ for the pattern-based methods.

\begin{table}[h!]
	\begin{subtable}{\linewidth}
		\centering
		\small
		\begin{tabular}{lrrrr}
			\toprule
			& \textsc{auc} & P & R & F\textsubscript{1} \\
			\midrule
			\multicolumn{5}{l}{\tiny RoBERTa-base}\\
			\textsc{nli} & 38.4 & 52.7 & 57.1 & \textbf{54.8} \\
			$\textsc{manpat}^{\pat\apat}$ & \textbf{46.1} & \textbf{64.0} & 45.4 & 53.1 \\
			$\textsc{manpat}^\pat$ & 32.4 & 32.4 & \textbf{94.5} & 48.2 \\
			$\textsc{autpat}_{5}^{\pat\apat}$ & 18.7 & 40.5 & 35.0 & 37.6 \\
			$\textsc{autpat}_{5}^{\pat}$ & 21.3 & 28.3 & 62.3 & 38.9 \\
			\midrule
			\multicolumn{5}{l}{\tiny RoBERTa-large}\\
			\textsc{nli} & 37.8 & 31.0 & 96.4 & 46.9 \\
			$\textsc{manpat}^{\pat\apat}$ & \textbf{70.4} & 39.6 & 95.3 & \textbf{56.0} \\
			$\textsc{manpat}^{\pat}$ & 38.9 & 25.6 & \textbf{98.3} & 40.6 \\
			$\textsc{autpat}_{5}^{\pat\apat}$ & 33.6 & \textbf{61.6} & 36.0 & 45.5 \\
			$\textsc{autpat}_{5}^{\pat}$ & 9.3 & 30.7 & 76.6 & 43.8 \\
			\bottomrule
		\end{tabular}
		\caption{SherLIiC train $\rightarrow$ Levy/Holt test.}
		\label{tab:transfer1}
	\end{subtable}
	
	\begin{subtable}{\linewidth}
		\centering
		\small
		\begin{tabular}{lrrrr}
			\toprule
			& \textsc{auc} & P & R & F\textsubscript{1} \\
			\midrule
			\multicolumn{5}{l}{\tiny RoBERTa-base}\\
			\textsc{nli} & 63.3 & 62.8 & \textbf{68.4} & \textbf{65.5} \\
			$\textsc{manpat}^{\pat\apat}$ & \textbf{69.1} & \textbf{80.5} & 42.1 & 55.3 \\
			$\textsc{manpat}^\pat$ & 68.4 & 80.1 & 24.2 & 37.2 \\
			$\textsc{autpat}_{5}^{\pat\apat}$ & 60.3 & 71.5 & 54.5 & 61.9 \\
			$\textsc{autpat}_{5}^{\pat}$ & 58.9 & 68.6 & 55.7 & 61.5 \\
			\midrule
			\multicolumn{5}{l}{\tiny RoBERTa-large}\\
			\textsc{nli} & 65.6 & 73.8 & 53.0 & 61.7 \\
			$\textsc{manpat}^{\pat\apat}$ & 69.6 & 84.7 & 35.7 & 50.3 \\
			$\textsc{manpat}^{\pat}$ & \textbf{72.2} & \textbf{89.3} & 30.3 & 45.2 \\
			$\textsc{autpat}_{5}^{\pat\apat}$ & 62.1 & 68.1 & \textbf{57.3} & \textbf{62.3} \\
			$\textsc{autpat}_{5}^{\pat}$ & 63.8 & 75.8 & 44.2 & 55.8 \\
			\bottomrule
		\end{tabular}
		\caption{Levy/Holt train $\rightarrow$ SherLIiC test.}
		\label{tab:transfer2}
	\end{subtable}
	\caption{Transfer learning. Table format: see \cref{tab:levyholt}.}
	\label{tab:transfer}
\end{table}

\begin{table*}
	\centering
	\begin{tabular}{lrrrrr}
		\toprule
		& \textsc{nli} & $\textsc{manpat}^{\pat\apat}$ & $\textsc{manpat}^{\pat}$ & $\textsc{autpat}_{5}^{\pat\apat}$ & $\textsc{autpat}_{5}^{\pat}$ \\
		\midrule
		$\mathrm{lr}$ & $2.72\cdot 10^{-5}$ & $2.47\cdot 10^{-5}$ & $6.68\cdot 10^{-6}$ & $3.82\cdot 10^{-5}$ & $2.11\cdot 10^{-5}$ \\
		$\lambda$ & $1.43\cdot 10^{-3}$ & $2.98\cdot 10^{-4}$ & $1.07\cdot 10^{-5}$ & $4.02\cdot 10^{-5}$ & $1.65\cdot 10^{-5}$ \\
		$c$ & $1$ & $2$ & $1$ & $2$ & $3$ \\
		\bottomrule
	\end{tabular}
	\caption{Levy/Holt; RoBERTa-base.}
	\label{tab:hparam_levyholt_base}
\end{table*}

\begin{table*}
	\centering
	\begin{tabular}{lrrrrrrr}
		\toprule
		& \textsc{nli} & $\textsc{manpat}^{\pat\apat}$ & $\textsc{manpat}^{\pat}$ & $\textsc{autpat}_5^{\pat\apat}$ & $\textsc{autpat}_5^{\pat}$ & $\textsc{autcur}_5^{\pat}$ & $\textsc{autarg}_5^{\pat}$ \\
		\midrule
		$\mathrm{lr}$ & $6.34\cdot 10^{-6}$ & $3.87\cdot 10^{-5}$ & $2.28\cdot 10^{-5}$ & $3.92\cdot 10^{-5}$ & $2.53 \cdot 10^{-5}$ & $1.28\cdot 10^{-5}$ & $2.47\cdot 10^{-5}$ \\
		$\lambda$ & $1.35\cdot 10^{-3}$ & $1.43\cdot 10^{-5}$ & $6.52\cdot 10^{-2}$ & $2.18\cdot 10^{-4}$ & $1.02\cdot 10^{-5}$ & $8.23\cdot 10^{-3}$ & $2.98\cdot 10^{-4}$ \\
		$c$ & $1$ & $4$ & $2$ & $1$ & $1$ & $1$ & $2$ \\
		\bottomrule
	\end{tabular}
	\caption{SherLIiC; RoBERTa-base.}
	\label{tab:hparam_sherliic_base}
\end{table*}

\begin{table*}
	\centering
	\begin{tabular}{lrrrrr}
		\toprule
		& \textsc{nli} & $\textsc{manpat}^{\pat\apat}$ & $\textsc{manpat}^{\pat}$ & $\textsc{autpat}_5^{\pat\apat}$ & $\textsc{autpat}_5^{\pat}$ \\
		\midrule
		$\mathrm{lr}$ & $6.68\cdot 10^{-6}$ & $4.55\cdot 10^{-6}$ & $4.92\cdot 10^{-6}$ & $6.68\cdot 10^{-6}$ & $8.13\cdot 10^{-6}$ \\
		$\lambda$ & $1.07\cdot 10^{-5}$ & $3.90 \cdot 10^{-4}$ & $3.61\cdot 10^{-4}$ & $1.07\cdot 10^{-5}$ & $6.05\cdot 10^{-2}$ \\
		$c$ & $1$ & $2$ & $3$ & $1$ & $2$ \\
		\bottomrule
	\end{tabular}
	\caption{Levy/Holt; RoBERTa-large.}
	\label{tab:hparam_levyholt_large}
\end{table*}

\begin{table*}
	\centering
	\begin{tabular}{lrrrrr}
		\toprule
		& \textsc{nli} & $\textsc{manpat}^{\pat\apat}$ & $\textsc{manpat}^{\pat}$ & $\textsc{autpat}_5^{\pat\apat}$ & $\textsc{autpat}_5^{\pat}$ \\
		\midrule
		$\mathrm{lr}$ & $6.68\cdot 10^{-6}$ & $1.29\cdot 10^{-5}$ & $9.14\cdot 10^{-6}$ & $6.34\cdot 10^{-6}$ & $4.55\cdot 10^{-6}$ \\
		$\lambda$ & $1.07\cdot 10^{-5}$ & $2.49\cdot 10^{-4}$ & $6.09\cdot 10^{-5}$ & $1.35\cdot 10^{-3}$ & $3.90\cdot 10^{-4}$ \\
		$c$ & $1$ & $3$ & $4$ & $1$ & $2$ \\
		\bottomrule
	\end{tabular}
	\caption{SherLIiC; RoBERTa-large.}
	\label{tab:hparam_sherliic_large}
\end{table*}

\begin{table*}
	\centering
	\small
	\begin{tabular}{l@{\, }lrrrrrrrrrrrr}
		\toprule
		&& \multicolumn{4}{c}{\dev{1}} & \multicolumn{4}{c}{\dev{2}} & \multicolumn{4}{c}{test}\\
		\cmidrule(lr){3-6}\cmidrule(lr){7-10}\cmidrule(lr){11-14}
		&& AUC & P & R & F1 & AUC & P & R & F1 & AUC & P & R & F1 \\
		\midrule
		{\tiny baselines}\\
		\multicolumn{2}{l}{\citet{hosseini-etal-2018-learning}} & -- & -- & -- & -- & -- & -- & -- & -- & 16.5 & -- & -- & -- \\  
		\multicolumn{2}{l}{\citet{hosseini-etal-2019-duality}} & -- & -- & -- & -- & -- & -- & -- & -- & \textbf{18.7} & -- & -- & -- \\
		\midrule
		\multicolumn{14}{l}{{\tiny RoBERTa-base}}\\
		\textsc{nli} & {\tiny ($\thr = \phantom{-}0.0052$)} & 94.9 & 87.4 & 91.1 & 89.2 & 88.8 & 78.1 & \textbf{90.3} & 83.8 & 72.6 & 68.7 & \textbf{75.3} & 71.9 \\
		$\textsc{manpat}^{\pat\apat}$ & {\tiny ($\thr = -0.0909$)} & \textbf{96.5} & \textbf{87.7} & \textbf{96.2} & \textbf{91.8} & \textbf{89.4} & \textbf{81.4} & 88.5 & \textbf{84.8} & \textbf{76.9} & \textbf{78.7} & 66.4 & \textbf{72.0} \\
		$\textsc{manpat}^{\pat}$ & {\tiny ($\thr = \phantom{-}0.5793$)} & 91.8 & 80.2 & 90.1 & 84.9 & 84.7 & 77.5 & 81.1 & 79.3 & 71.2 & 74.4 & 61.2 & 67.1 \\
		$\textsc{autpat}_{5}^{\pat\apat}$ & {\tiny ($\thr = -0.1428$)} & 95.0 & 83.4 & 95.6 & 89.1 & 87.7 & 79.2 & 85.7 & 82.3 & 63.7 & 71.0 & 58.8 & 64.3 \\
		$\textsc{autpat}_{5}^{\pat}$ & {\tiny ($\thr = -0.0592$)} & 87.4 & 78.0 & 90.0 & 83.6 & 83.3 & 76.3 & 81.6 & 78.8 & 65.4 & 68.0 & 63.3 & 65.5 \\
		\midrule
		\multicolumn{14}{l}{{\tiny RoBERTa-large}}\\
		\textsc{nli} & {\tiny ($\thr=\phantom{-}0.0016$)} & 96.9 & 90.1 & 97.1 & \textbf{93.5} & 87.7 & 82.6 & 87.6 & \textbf{85.0} & 75.5 & 73.5 & 73.7 & 73.6 \\
		$\textsc{manpat}^{\pat\apat}$ & {\tiny ($\thr = \phantom{-}0.1156$)} & \textbf{97.1} & \textbf{91.4} & 95.4 & 93.4 & \textbf{88.9} & \textbf{84.0} & 84.8 & 84.4 & \textbf{83.9} & \textbf{84.8} & 70.1 & \textbf{76.7} \\
		$\textsc{manpat}^{\pat}$ & {\tiny ($\thr = -0.8457$)} & 92.2 & 76.1 & \textbf{97.3} & 85.4 & 84.4 & 72.5 & \textbf{91.2} & 80.8 & 77.8 & 67.9 & \textbf{81.5} & 74.1 \\
		$\textsc{autpat}_{5}^{\pat\apat}$ & {\tiny ($\thr = -0.0021$)} & 95.0 & 86.0 & 91.9 & 88.8 & 84.7 & 78.9 & 81.1 & 80.0 & 70.4 & 75.7 & 60.7 & 67.4 \\
		$\textsc{autpat}_{5}^{\pat}$ & {\tiny ($\thr = -0.9197$)} & 92.4 & 75.5 & 95.3 & 84.2 & 83.5 & 70.6 & 88.5 & 78.5 & 66.5 & 61.8 & 74.4 & 67.5 \\
		\bottomrule
	\end{tabular}
	\caption{Full results on the Levy/Holt dataset. The dev and test sets as created by \citet{hosseini-etal-2018-learning} are called \dev{1} and test. The portion of \dev{1} that serves as our validation set is called \dev{2}. AUC denotes the area under the precision-recall curve for precision values${}\geq0.5$. All results in \%{}.}
	\label{tab:dev_levyholt}
\end{table*}

\begin{table*}
	\centering
	\small
	\begin{tabular}{l@{\, }lrrrrrrrrrrrr}
		\toprule
		&& \multicolumn{4}{c}{\dev{1}} & \multicolumn{4}{c}{\dev{2}} & \multicolumn{4}{c}{test}\\
		\cmidrule(lr){3-6}\cmidrule(lr){7-10}\cmidrule(lr){11-14}
		&& \textsc{auc} & P & R & F\textsubscript{1} & \textsc{auc} & P & R & F\textsubscript{1} & \textsc{auc} & P & R & F\textsubscript{1} \\
		\midrule
		\multicolumn{14}{l}{\tiny baselines}\\
		\multicolumn{2}{l}{Lemma} & -- & \textbf{90.0} & 10.9 & 19.4 & -- & -- & -- & -- & -- & \textbf{90.7} & 8.9 & 16.1 \\
		\multicolumn{2}{l}{w2v+untyped\_{}rel} & -- & 56.5 & 74.0 & 64.1 & -- & -- & -- & -- & -- & 52.8 & 69.5 & 60.0 \\
		\multicolumn{2}{l}{w2v+tsg\_rel\_emb} & -- & 56.6 & \textbf{77.6} & \textbf{65.5} & -- & -- & -- & -- & -- & 51.8 & \textbf{72.7} & \textbf{60.5} \\
		\midrule
		\multicolumn{14}{l}{\tiny RoBERTa-base}\\
		\textsc{nli} & {\tiny ($\thr = \phantom{-}0.3878$)} & 81.3 & \textbf{79.1} & 80.1 & 79.6 & 81.5 & \textbf{84.2} & 70.6 & \textbf{76.8} & 65.8 & \textbf{67.0} & 66.1 & 66.5 \\
		$\textsc{manpat}^{\pat\apat}$ & {\tiny ($\thr = -0.3324$)} & 76.2 & 68.6 & 85.8 & 76.2 & \textbf{82.4} & 70.0 & \textbf{82.4} & 75.7 & 66.4 & 60.9 & 78.8 & 68.7 \\
		$\textsc{manpat}^\pat$ & {\tiny ($\thr = -0.4812$)} & \textbf{88.4} & 75.5 & \textbf{93.1} & \textbf{83.4} & 84.1 & 73.0 & 79.4 & 76.1 & \textbf{69.2} & 62.0 & \textbf{81.2} & \textbf{70.3} \\
		$\textsc{autpat}_{5}^{\pat\apat}$ & {\tiny ($\thr = -0.4694$)} & 87.0 & 77.8 & 88.8 & 82.9 & 71.2 & 68.4 & 76.5 & 72.2 & 67.4 & 61.8 & 75.6 & 68.0 \\
		$\textsc{autpat}_{5}^{\pat}$ & {\tiny ($\thr = -0.7042$)} & 86.8 & 64.1 & 91.8 & 75.5 & 74.0 & 65.5 & 83.8 & 73.6 & 67.3 & 56.6 & 82.6 & 67.2 \\
		$\textsc{autcur}_{5}^{\pat}$ & {\tiny ($\thr = -0.7524$)} & 82.6 & 61.7 & 92.8 & 74.1 & 75.6 & 60.6 & 88.2 & 71.9 & 69.5 & 56.3 & 89.6 & 69.2 \\
		$\textsc{autarg}_{5}^{\pat}$ & {\tiny ($\thr = -0.7461$)} & 77.4 & 69.3 & 84.0 & 76.0 & 73.6 & 68.9 & 75.0 & 71.8 & 65.2 & 61.9 & 75.6 & 68.1 \\
		\midrule
		$\textsc{autpat}_{15}^{\pat\apat}$ & {\tiny ($\thr = -0.5263$)} & 95.3 & 87.0 & 93.1 & 89.9 & 73.0 & 65.4 & 75.0 & 69.9 & 70.0 & 60.4 & 79.7 & 68.7 \\
		$\textsc{autpat}_{15}^{\pat}$ & {\tiny ($\thr = -0.6422$)} & \textbf{95.4} & 85.3 & 94.6 & 89.7 & \textbf{75.8} & 69.2 & \textbf{79.4} & \textbf{74.0} & \textbf{73.1} & \textbf{63.0} & 77.4 & \textbf{69.4} \\
		$\textsc{autpat}_{25}^{\pat\apat}$ & {\tiny ($\thr = -0.0014$)} & 95.0 & \textbf{92.0} & 93.7 & \textbf{92.8} & 66.1 & \textbf{70.0} & 72.1 & 71.0 & 63.5 & 62.1 & 73.4 & 67.3 \\
		$\textsc{autpat}_{25}^{\pat}$ & {\tiny ($\thr = -0.6496$)} & 88.1 & 72.3 & 90.0 & 80.2 & 73.0 & 67.5 & 79.4 & 73.0 & 69.0 & 60.5 & \textbf{79.4} & 68.7 \\
		$\textsc{autpat}_{50}^{\pat\apat}$ & {\tiny ($\thr = -0.9163$)} & 93.2 & 72.8 & 92.8 & 81.5 & 67.1 & 63.0 & 75.0 & 68.5 & 66.3 & 54.3 & \textbf{82.8} & 65.6 \\
		$\textsc{autpat}_{50}^{\pat}$ & {\tiny ($\thr = -0.9500$)} & 94.2 & 79.1 & \textbf{94.9} & 86.3 & 69.3 & 66.3 & 77.9 & 71.6 & 67.4 & 57.3 & 82.5 & 67.6 \\
		\midrule
		\multicolumn{14}{l}{\tiny RoBERTa-large}\\
		\textsc{nli} & {\tiny ($\thr = \phantom{-}0.0025$)} & \textbf{92.3} & \textbf{79.7} & \textbf{93.7} & \textbf{86.1} & 75.7 & 66.7 & \textbf{82.4} & 73.7 & 68.3 & 60.5 & \textbf{85.5} & 70.9 \\
		$\textsc{manpat}^{\pat\apat}$ & {\tiny ($\thr = -0.0956$)} & 89.3 & 77.3 & 88.5 & 82.5 & \textbf{80.8} & \textbf{74.7} & 77.9 & \textbf{76.3} & \textbf{74.4} & \textbf{66.0} & 80.8 & \textbf{72.6} \\
		$\textsc{manpat}^{\pat}$ & {\tiny ($\thr = -0.6641$)} & 78.0 & 67.4 & 84.9 & 75.1 & 72.2 & 67.1 & 77.9 & 72.1 & 64.6 & 58.1 & 79.0 & 67.0 \\
		$\textsc{autpat}_{5}^{\pat\apat}$ & {\tiny ($\thr = -0.9889$)} & 86.5 & 73.8 & 86.7 & 79.7 & 73.6 & 70.4 & 73.5 & 71.9 & 68.6 & 61.9 & 75.5 & 68.0 \\
		$\textsc{autpat}_{5}^{\pat}$ & {\tiny ($\thr = -0.5355$)} & 71.6 & 64.3 & 71.9 & 67.9 & 64.5 & 71.4 & 66.2 & 68.7 & 56.8 & 61.5 & 66.1 & 63.7 \\
		\bottomrule
	\end{tabular}
	\caption{Full results on SherLIiC. The original dev and test sets are called \dev{1} and test. The portion of \dev{1} that serves as our validation set is called \dev{2}. \textsc{auc} denotes the area under the precision-recall curve for precision values${}\geq0.5$. Baseline results from \citep{schmitt-schutze-2019-sherliic}. All results in \%{}.}
	\label{tab:dev_sherliic}
\end{table*}

\begin{table*}
	\centering
	\small
	\begin{tabularx}{\linewidth}{lXcX}
		\toprule
		(1) &The original aim of de Garis' work was to \textbf{prem} the field of "brain building" (a term of his invention) and to "\textbf{hypo} a trillion dollar industry within 20 years". &$\to$& The original aim of their work was that "\textsc{pargl} \textbf{prem} \textsc{pargr}" and that "\textsc{hargl} \textbf{hypo} \textsc{hargr} within 20 years". \\
		(2) &Critic Roger Ebert stated that Gellar and co-star Ryan Phillippe "\textbf{prem} a convincing emotional charge" and that Gellar is "effective as a bright girl who knows exactly how to \textbf{hypo} her act as a tramp". &$\to$& Critic Roger Ebert stated that \textsc{pargl} and co-star Ryan Phillippe "\textbf{prem} \textsc{pargr}" and that \textsc{hargl} is "effective as a bright girl who knows exactly how she \textbf{hypo} \textsc{hargr} as a tramp".\\
		(3) &Well-known professional competitions in the past have included the World Professional Championships (\textbf{hypo} Landover, Maryland), the Challenge Of Champions, the Canadian Professional Championships and the World Professional Championships (\textbf{prem} in Jaca, Spain). &$\to$& Well-known professional competitions in the past have included \textsc{hargl} (\textbf{hypo} \textsc{hargr}), the  Challenge Of Champions, the Canadian Professional Championships and \textsc{pargl} (\textbf{prem} \textsc{pargr}).\\
		(4) &They also had sharpshooter Steve Kerr, whom they \textbf{hypo} via free agency before the 1993–94 season, Myers, and centers Luc Longley (\textbf{prem} via trade in 1994 from the Minnesota Timberwolves) and Bill Wennington. &$\to$& \textsc{hargl} also had sharpshooter \textsc{hargr}, whom they \textbf{hypo} via free agency before the 1993–94 season, Myers, and centers \textsc{pargr} (whom \textsc{pargl} \textbf{prem} via trade in 1994 from the Minnesota Timberwolves) and Bill Wennington.\\
		(5) &Because the 6x86 was more efficient on an instructions-per-cycle basis than Intel's Pentium, and because Cyrix sometimes \textbf{hypo} a faster bus speed than either Intel or AMD, Cyrix and competitor AMD co-\textbf{prem} the controversial PR system in an effort to compare its products more favorably with Intel's. \dots &$\to$& Because the 6x86 was more efficient on an instructions-per-cycle basis than Intel's Pentium, and because \textsc{hargl} sometimes \textbf{hypo} \textsc{hargr}, \textsc{pargl} and competitor AMD co-\textbf{prem} \textsc{pargr} in an effort to compare its products more favorably with Intel's. \dots \\
		\bottomrule
	\end{tabularx}
	\caption{Five manually selected patterns from the 100 highest-ranked automatically extracted patterns from SherLIiC \dev{1} (used in $\textsc{autcur}^{\Phi}_{5}$) and their rewritten counterparts (used in $\textsc{autarg}^{\Phi}_{5}$). \textsc{pargl} (\textsc{hargl}) stands for the left argument of the premise (hypothesis); \textsc{pargr} (\textsc{hargr}) for the right one.}
	\label{tab:pattern_rewriting}
\end{table*}

\end{document}